\documentclass[11pt]{article}

\usepackage[preprint]{acl}

\usepackage{times}
\usepackage{latexsym}
\usepackage[T1]{fontenc}
\usepackage[utf8]{inputenc}
\usepackage{microtype}
\usepackage{inconsolata}
\usepackage{graphicx}
\usepackage{booktabs}
\usepackage{colortbl}% row shading in the main table (xcolor is loaded by acl.sty)
\usepackage{amsmath}
\usepackage{amssymb}
\usepackage{multirow}
\usepackage{array}
\usepackage{xspace}
\usepackage{pifont}
\usepackage{fvextra}% Verbatim with line wrapping (appendix prompt boxes); loads fancyvrb
\usepackage[skins,breakable]{tcolorbox}

\definecolor{feInk}{HTML}{1E1E1E}
\definecolor{feBlue}{HTML}{8AABCA}
\definecolor{feBlueDark}{HTML}{5B86AD}
\definecolor{feBrown}{HTML}{846358}
\definecolor{feLight}{HTML}{E1EBF5}
\definecolor{feGrey}{HTML}{9A9A9A}
\tcbset{fecard/.style={enhanced, arc=1.5pt, outer arc=1.5pt,
  boxrule=0.6pt, colframe=feBlueDark, colback=feLight!45!white, coltext=feInk,
  coltitle=white, colbacktitle=feBlueDark,
  fonttitle=\sffamily\bfseries\scriptsize,
  toptitle=0.5mm, bottomtitle=0.5mm, lefttitle=4pt,
  left=4pt, right=4pt, top=3pt, bottom=3pt,
  before skip=6pt plus 2pt, after skip=4pt plus 2pt}}
\newtcolorbox{promptbox}[2][]{fecard, breakable, title={#2}, #1}
\newtcolorbox{codebox}[2][]{fecard, title={#2}, #1}
\newtcolorbox{schemabox}[2][]{fecard, breakable, colframe=feBrown, colbacktitle=feBrown,
  colback=feLight!30!white, before skip=4pt plus 2pt, after skip=4pt plus 2pt,
  fontupper=\ttfamily\footnotesize\raggedright, title={#2}, #1}
\DefineVerbatimEnvironment{PromptVerbatim}{Verbatim}{fontsize=\scriptsize,
  breaklines=true, breaksymbolleft={}, breakindent=0pt}
\DefineVerbatimEnvironment{DiffVerbatim}{Verbatim}{fontsize=\scriptsize}

\newcommand{\method}{LegoEnv\xspace}
\newcommand{\bench}{AutomationBench\xspace}

\newcommand{\basea}{Qwen3.5-27B\xspace}   % weakest base,  1.3% zero-shot
\newcommand{\baseb}{Qwen3.6-27B\xspace}   % mid base,      8.0% zero-shot
\newcommand{\basec}{Qwen3.8-27B\xspace}   % strongest base, 23.3%; also the teacher

\newcommand{\cmark}{\ding{51}}
\newcommand{\xmark}{\ding{55}}

\newcommand{\todo}[1]{}% leader-review build; restore: \textcolor{red}{[TODO: #1]}

\newcommand{\ie}{\emph{i.e., }}

\title{\method: Self-Evolving Agentic Environments\\via Modular Recombination}

\author{Zhuosong Jiang$^{1*\diamond}$, Chengxu Wu$^{2*\diamond}$, Junqi Liu$^{3*\diamond}$, Yujie Yi$^{4*\diamond}$, Siqin Wang$^{1}$, \\
\bfseries Zhixuan Huang$^{2\diamond}$, Jinqian Jiang$^{2\diamond}$, Lijun Xiao$^{1}$, Jianxun Li$^{5}$, Yongyang Pan$^{5}$, \\
\bfseries Jieran Wang$^{5}$, Shiqi Shen$^{5}$, Fei Liu$^{6}$, Xinyu Lin$^{1}$ \\[4pt]
{\small\mdseries $^{1}$National University of Singapore \quad $^{2}$Peking University \quad $^{3}$University of California, Santa Cruz} \\
{\small\mdseries $^{4}$Shanghai Jiao Tong University \quad $^{5}$Tencent \quad $^{6}$Emory University}}

\begin{document}
\maketitle
{\let\thefootnote\relax\footnotetext{$^{*}$Equal contribution. $^{\diamond}$Work done during an internship at Tencent. Code: \url{https://github.com/JazyJiang/LegoEnv}. Correspondence: \texttt{zhuosong.jiang@u.nus.edu}.}}

\begin{abstract}
LLM agents learn by interacting with environments, yet existing environments are designed for specific tasks and fail to generalize to new ones.
Recent methods automate environment construction with LLMs.
However, they generate each environment from scratch, neither reusing parts shared across environments nor carrying a fix from one environment to the rest.
To address this, we propose a new paradigm of \emph{evolving environment atoms}, which composes environments from reusable atoms (intent, rule, data and server) and evolves the atoms when an environment fails to give reliable feedback (e.g., an application lacks a tool the task requires).
To instantiate this paradigm, we introduce \method, which builds 5{,}465 verified environments over 223 applications, and one round of evolution raises the share of solvable environments from 71.6\% to 88.3\%.
% On \bench, supervised fine-tuning on these environments lifts \baseb from 16.0\% to 32.0\%, and reinforcement learning with their verifiers moves \basec from 37.6\% to 40.3\%.
% Further analyses show that performance grows with the corpus and transfers to unseen applications, and that failures trace back to a few shared root causes.
On \bench, SFT on these environments lifts \baseb from 16.0\% to 32.0\%, and RL with their verifiers lifts \basec from 37.6\% to 40.3\%. Further analyses show steady gains with corpus size, transfer to unseen applications, and failures concentrated in a few shared root causes.
\end{abstract}

\section{Introduction}
\label{sec:intro}

LLM agents have demonstrated remarkable performance across a wide range of domains, spanning from resolving complex issues in software engineering~\citep{swebench, sweagent} and solving competition mathematics~\citep{deepseekr1} to discovering new knowledge and algorithms~\citep{aiscientist, alphaevolve}.
Agents acquire these abilities by interacting with environments, where the agent acts, the environment responds, and the response becomes supervision, as trajectories for supervised fine-tuning~\citep{agenttuning, apigenmt, toucan, swegym} or rewards for reinforcement learning~\citep{deepseekr1, toolrl, retool, arpo, verltool, agentrlvr}.
Since an agent can learn only what its environments can pose and check, the construction and quality of environments are among the most critical steps in agent training.

\begin{figure*}[t]
\centering
\includegraphics[width=0.95\textwidth]{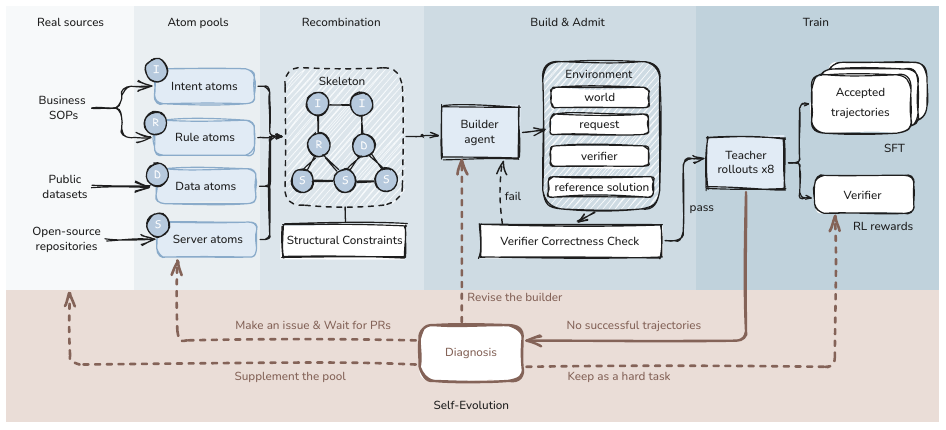}% matplotlib version: figures/overview.pdf
\caption{Overview of \method. Four pools of atoms are collected from real sources and composed into skeletons under structural constraints. A builder agent turns each skeleton into an environment, which is admitted only after its verifier passes a correctness check. Environments that yield no successful trajectory are diagnosed, and the result files an issue against an atom and waits for its fix, supplements a pool, or revises the builder; tasks that are merely hard are kept.}
\label{fig:overview}
\end{figure*}

% \xinyu{P2. Issue: task-specific environments fail to generalize. Existing environments are designed per task (data, tools, checker) for high-quality feedback; footnote defines one environment as one concrete task $\rightarrow$ example: an invoice-payment environment $\rightarrow$ when the task becomes harder or moves to a new scenario, the environment fails $\rightarrow$ reason: tasks differ in data, tools, operations and checks $\rightarrow$ scalable environment construction is essential.}
Existing environments are typically designed for a specific task, with the data, the tools and the verifier built together to provide high-quality feedback~\citep{webarena, osworld, taubench, appworld, automationbench, mcpmark}.\footnote{We count one environment per concrete task, together with the applications, data and verifier it requires. For example, paying the approved invoices of a month is one environment, and escalating the overdue invoices on the same applications is another.}
However, they fail to generalize when the task becomes more complex or moves to a new scenario (e.g., an office agent's task shifts from paying invoices in an accounting system to onboarding employees in an HR system), since different tasks require different data, tools, operations and checks.
Therefore, scalable construction that quickly builds high-quality environments for diverse tasks is essential for training agents effectively and efficiently.

% \xinyu{P3. Gap: scalable-environment methods generate each environment from scratch. LLMs generate tools, data and tasks per environment $\rightarrow$ but environments share common parts and differ in task-specific ones $\rightarrow$ generating from scratch cannot reuse previous construction experience and wastes cost on rebuilding shared parts.}
To scale environments up, recent methods use LLMs to generate the tools, data and tasks of each environment~\citep{agentscaler, envscaler, scaleenv, awm, envfactory, envcraft, agentworld}.
However, different environments share common parts (e.g., a mail application for sign-off and an approval policy) while differing in task-specific ones (e.g., invoice records versus employee profiles).
Generating each environment from scratch thus neither reuses the experience gained from previous constructions nor avoids the cost of rebuilding the shared parts (Table~\ref{tab:related}).

% \xinyu{P4. Our paradigm: evolving atoms for scalable agentic environments. Decompose an environment into four atoms (intent, rule, data, server) $\rightarrow$ compose atoms into environments, evolve atoms with failures $\rightarrow$ benefits: scalable, self-evolving, grounded in real-world needs.}
Based on the above insights, we propose a new paradigm, \emph{evolving atoms} for scalable agentic environments, which builds environments from smaller reusable units.
Specifically, we decompose an environment into four kinds of atoms, namely \textit{intent} (goal to achieve), \textit{rule} (business rule to follow), \textit{data} (records for the agent to use), and \textit{server} (tool provider), and construct each environment by composing them.
When an environment fails, we trace the failure to the atoms behind it and evolve those atoms, so that every later environment built from them benefits.
This paradigm brings three benefits.
\textbf{(1) Scalability.} New environments are created by composing existing atoms, so their number grows with the compositions rather than with manual effort.
% \textbf{(2) Self-evolution.} The atoms expand wherever failures expose gaps, so every later environment inherits the fix.
\textbf{(2) Self-evolution.} Atom sets and atom properties will be evolved from construction failures, therefore allowing later environments to avoid the same errors and reuse the new atoms. 
\textbf{(3) Real-world grounding.} Every atom is initially collected from a real source, so the environments reflect real-world needs and carry practical value. 

% \xinyu{P5. Framework and results. Collect atoms from real sources $\rightarrow$ recombine them under structural constraints $\rightarrow$ admit environments with tested verifiers $\rightarrow$ diagnose failures to evolve atoms $\rightarrow$ trained agents improve on three benchmarks and beat SOTA pipelines.}
% (10-10: adopted, \method is now LegoEnv) \xinyu{Would it be good to name the method as ``LegoEnv''? As Lego is very intuitive for readers to understand that we modularize the environments into different atoms.}
Based on this paradigm, we develop \method (Figure~\ref{fig:overview}) for business domains.
\method first collects atoms from real sources and recombines them under structural constraints.
% It then builds an environment from each valid composition and admits the environment only after its verifier passes a correctness test.
It then builds an environment from each valid composition and keeps it only if its verifier passes a correctness test. 
% Finally, it diagnoses the environments that no teacher can solve and evolves the atoms behind them, which raises the share of solvable environments from 71.6\% to 88.3\% in one round.
Finally, it diagnoses the environments that no teacher can solve and evolves the atoms that cause the failures, which raises the share of solvable environments from 71.6\% to 88.3\% in one round. 
% We evaluate \method on \bench~\citep{automationbench}, MCP-Mark~\citep{mcpmark} and BFCL v4~\citep{bfcl}.
% Supervised fine-tuning on our environments lifts \baseb from 16.0\% to 32.0\% and \basea from 3.2\% to 23.2\% on \bench; under one teacher, student and trajectory budget our environments exceed four public synthesis pipelines by 8.0 to 13.7 points.
% Moreover, using our verifiers as rewards for reinforcement learning moves \basec from 37.6\% to 40.3\%.
We evaluate \method on \bench~\citep{automationbench}, MCP-Mark~\citep{mcpmark} and BFCL v4~\citep{bfcl}.
Supervised fine-tuning on our environments improves \baseb by 100\% and \basea by 625\% on \bench, outperforming four public synthesis pipelines by 37\% to 87\% under the same budget.
Using the environments' verifiers as rewards for reinforcement learning further improves \basec by 7.2\%.

Our contributions are threefold:
\begin{list}{\labelitemi}{\leftmargin=1.2em\topsep=2pt\itemsep=1pt\parsep=0pt\partopsep=0pt}
\item We propose \emph{evolving atoms}, a paradigm that composes environments from reusable intent, rule, data, and server atoms, which are collected from real sources and evolve over time.
\item We instantiate it as  \method, which tests every verifier before use and attributes each environment fault to an atom, a verifier kind or a builder instruction, so that one repair reaches every environment that shares it.
\item We show that the environments improve three bases on \bench, with supervised gains that transfer to MCP-Mark, exceed four public pipelines at equal budget, and serve unchanged as rewards for reinforcement learning.
\end{list}
% P6 Contributions: dropped 10-10 for space, restored 10-12.

\section{Related Work}
\label{sec:related}

% \paragraph{Environment Construction.} Existing pipelines synthesise environments from tools and generated content (Table~\ref{tab:related}, Appendix~\ref{app:related}): \citet{agentscaler} sample tool sequences through graph walks and \citet{envfactory} guide query generation with dependency-aware sampling; \citet{envscaler} seed fictitious data, \citet{awm} and \citet{scaleenv} generate records from schemas; servers come from API definitions~\citep{agentscaler}, website names~\citep{awm} or domain names~\citep{scaleenv}. Scalable synthesis has also been pursued for web~\citep{autowebworld, infiniteweb, webworld}, computer-use~\citep{cuagym, opencomputer}, terminal~\citep{litecoderterminal, metatask} and software-engineering~\citep{menvagent} environments, and at ecosystem scale~\citep{nexn1}. Closest to us, \citet{compoworld} compose tasks from a finite library of services with crawled interfaces and generated implementations, whereas our atoms also cover intents, rules and data, and change when environments fail. \method draws intents from automation templates and procedures, data from public datasets and servers from open-source applications; it places published rules inside the world for agents to discover, whereas \citet{cworld} and \citet{envscaler} put them in the instruction or the documentation; and it seeds the applications of one task from one dataset so that records correspond across them (\S\ref{sec:admission}), whereas \citet{envfactory}, \citet{cworld} and \citet{agentskiller} link applications by parameter similarity, retrieval or database merging.
\paragraph{Environment Construction.} 
Building high-quality environments is an essential step for agent learning.  
One line of work builds scalable environments for specific scenarios, such as web~\citep{autowebworld, infiniteweb, webworld}, computer use~\citep{cuagym, opencomputer}, terminal~\citep{litecoderterminal, metatask} and software engineering~\citep{menvagent}, or within one ecosystem~\citep{nexn1}, so each new scenario requires redesigning the synthesis rules by hand.
To generalize across scenarios, recent pipelines automate construction with LLMs: they generate servers from API definitions~\citep{agentscaler}, website names~\citep{awm} or domain names~\citep{scaleenv}, records from schemas or fictitious seeds~\citep{envscaler, awm, scaleenv}, and tasks from tool graphs or dependencies~\citep{agentscaler, envfactory}, and they link applications by parameter similarity, retrieval or database merging~\citep{envfactory, cworld, agentskiller}.
However, they generate each environment from scratch, so shared parts are rebuilt every time and a fix in one environment never reaches the others.
To enable automatic environment construction across tasks, \citet{compoworld} reuse a library of services, but the library remains fixed when generating environments. 
In contrast, \method composes environments from reusable intent, rule, data and server atoms collected from real sources, and evolves these atoms when environments fail, so every later environment inherits the fix (Table~\ref{tab:related}, Appendix~\ref{app:related}). 

% \paragraph{Environment Verification.} Success is decided with an LLM judge, alone or beside rules~\citep{awm, agentworld, cworld, covert, toucan}, or by comparing final states~\citep{agentscaler, scaleenv, envfactory, autoforge}; both rely on the verifier being correct. \citet{trustworthyworlds} repair synthetic web environments to make tasks solvable. Our admission test checks that each verifier distinguishes successful execution from failure before the environment is used.
\paragraph{Environment Verification.}
Existing pipelines decide whether an agent completes a task with an LLM judge, alone or with rules~\citep{awm, agentworld, cworld, covert, toucan}, or by comparing final states~\citep{agentscaler, scaleenv, envfactory, autoforge}, assuming the verifier itself is correct; a wrong verifier silently rewards wrong behavior.
\citet{trustworthyworlds} repair synthetic web environments, but only to make tasks solvable.
In contrast, \method tests each verifier before use: it must fail on the initial world and pass on the reference solution, or its environment is discarded.

% \paragraph{Environment Evolution.} \citet{agentworld} generate targeted tasks where the current policy fails, \citet{genenv} and \citet{spade} co-evolve a task generator with the learner, and \citet{envcraft} and \citet{trustworthyworlds} repair the individual environments that fail their audits. We trace teacher failures to verifier defects, world defects or task difficulty, and repair the shared atoms or the composition vocabulary, so later environments inherit the fix (\S\ref{sec:evolve}).
\paragraph{Environment Evolution.}
To constantly give reliable feedback across tasks on different agents, environment evolution has become a promising solution. 
Some work evolves what is asked: \citet{agentworld} generate tasks where the current policy fails, and \citet{genenv} and \citet{spade} co-evolve a task generator with the learner.
Alternatively, some work fixes errors by repairing each failed environment individually~\citep{envcraft, trustworthyworlds}.
In contrast, \method traces each failure to a verifier defect, a world defect, or task difficulty, and repairs the shared atoms or composition vocabulary behind it, so every environment built from them inherits the fix (\S\ref{sec:evolve}).

\section{Method}
\label{sec:method}

\begin{figure}[t]
\centering
\includegraphics[width=0.92\columnwidth]{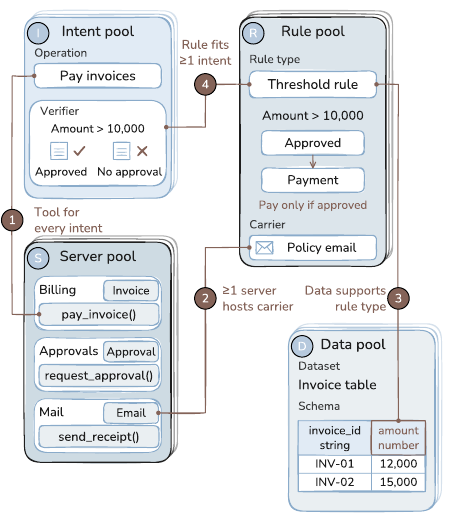}
\caption{The four atom pools on one worked example (pay invoices over 10{,}000 only if approved). The brown links, numbered 1--4, are the four structural constraints of \S\ref{sec:recombine}.}
\label{fig:atoms}
\end{figure}

% \method builds an environment in three steps and asks one question at each. Atoms are composed into a skeleton: do these atoms fit together (\S\ref{sec:recombine})? A builder turns the skeleton into an environment: can its verifier be trusted (\S\ref{sec:admission})? A teacher attempts the task: can a capable agent do it at all (\S\ref{sec:evolve})? 
% The questions cost more in that order, and each no means something different. 
% If the atoms do not fit, the skeleton is dropped. 
% If the verifier cannot be trusted, the builder revises. 
% If no teacher attempt succeeds, the environment is diagnosed, and the diagnosis repairs an atom, revises the builder, supplements a pool, or keeps the task as a hard one. 
In this section, we first introduce two key basic components in \method, \ie shared condition form (\S\ref{sec:grammar}) for environment evolution; and atoms (\S\ref{sec:atoms}), the basic unit for environment construction. 
Based on the components, \method builds each environment through three progressive checks.
First, atoms are composed into a skeleton, which is dropped if the atoms are incompatible (\S\ref{sec:recombine}).
Second, a builder turns the skeleton into an environment, and revises it until its verifier passes a correctness test (\S\ref{sec:admission}).
Third, a teacher attempts the task; if every attempt fails, a diagnosis agent finds the cause and either repairs an atom, revises the builder, supplements a pool, or keeps the task as a hard one (\S\ref{sec:evolve}). 
Figure~\ref{fig:overview} shows the loop and Figure~\ref{fig:atoms} illustrates the atoms and the constraints between them.

% Table of atom families, replaced by Figure~\ref{fig:atoms}; restore if wanted.
% \begin{table}[t]
% \centering
% \small
% \setlength{\tabcolsep}{3pt}
% \begin{tabular}{@{}lp{0.40\columnwidth}p{0.36\columnwidth}@{}}
% \toprule
% Atom & Holds & Pool \\
% \midrule
% intent  & server, operation kind, required or forbidden, verifier kinds & 110 operation kinds, 45 verifier kinds \\
% rule    & archetype, carrier, exception, shape & 8 archetypes, 6 carriers \\
% data    & public dataset and its card & 281 usable of 336 \\
% server & entities, tools, preconditions & 384 compiled, 223 used \\
% \bottomrule
% \end{tabular}
% \caption{The four atom families.}
% \label{tab:atoms}
% \end{table}

\subsection{Shared condition form}
\label{sec:grammar} 
To build LegoEnv and achieve the three progressive checks, we first introduce a shared condition form. 
A condition is a triple $\langle e,\varphi,\kappa\rangle$: an entity $e$, a scope $\varphi$, which is a conjunction of comparisons on the columns of $e$, and a cardinality $\kappa$. It holds when the number of rows of $e$ that satisfy $\varphi$ meets $\kappa$. For example, a server lets an invoice be paid only if it has been approved: {\small $e=\mathtt{Approval}$, $\varphi=\{\mathtt{invoice}=\mathtt{\$arg.id},\ \mathtt{status}=\mathtt{approved}\}$, $\kappa\geq 1$}. A verifier asserts, in the same form, that no overdue payable was left unescalated: {\small $e=\mathtt{Payable}$, $\varphi=\{\mathtt{due}<\mathtt{\$now},\ \mathtt{escalated}=\mathtt{false}\}$, $\kappa=0$}. The conditions a server checks before running a tool, the constraints on its tables, and the assertions of a verifier are all written in this one form and evaluated by one evaluator. A verifier can therefore say exactly what a server can check, no more and no less. Appendix~\ref{app:grammar} gives the full grammar.

\subsection{Atoms}
\label{sec:atoms}

Each atom is collected once from a real source and reused. Appendix~\ref{app:atoms} gives the schema and a real instance of each.

\paragraph{Server atoms.} A server atom is one application, compiled from its open-source code: its entities, the tools that operate on them, the condition each tool checks before it runs, and the message the tool returns when the condition fails, which is what the agent sees when it hits a wall. From 467 open-source repositories we obtain 384 usable servers (Appendix~\ref{app:compile}); the environments in this paper use 223 of them.

\paragraph{Data atoms.} A data atom is a public dataset with a card that records which tables have enough rows, the type of each column, and which kinds of rule the columns can support: a threshold needs a numeric column, a deadline a date. Worlds are seeded from these datasets, never from generated records. Of 336 datasets, 281 have a usable table.

\paragraph{Intent atoms.} An intent atom specifies a single write that must, or must not, happen on a server. It is defined by an \emph{operation kind}, a recurring class of write across applications such as \emph{create invoice} or \emph{post message}; each server's card says which of its tools implement which kinds. Whether an intent was fulfilled is checked by one or more \emph{verifier kinds}, templates over the condition form of \S\ref{sec:grammar} such as \emph{mail sent with terms} or \emph{row cell equals}; the verifier kinds were grounded in public documentation of business workflows (Appendix~\ref{app:verifierkinds}). Across our corpus, the vocabulary covers 110 operation kinds and 45 verifier kinds. The vocabulary is not fixed: it expands whenever new servers introduce write operations that no kind yet expresses (\S\ref{sec:evolve}). An operation kind can \emph{carry a rule} if its verifier kinds can tell valid records from invalid ones.

\paragraph{Rule atoms.} A rule atom is a business rule the agent has to find and follow: an \emph{archetype}, the kind of rule it is (threshold routing, exclusion, deadline escalation, tiering, approval chain, budget quota, reconciliation, compliance flag), a shape free of names and numbers, an \emph{optional exception} that overrides the rule for some records (\emph{unless marked urgent}, \emph{unless approved by a manager}), and a \emph{carrier}, the place in the world that holds the rule's text, such as a policy worksheet, a document, a mail or a note on a record. The request names the carrier and never restates the rule.

\subsection{Structural atom recombination}
\label{sec:recombine}

\begin{figure}[t]
\centering
\includegraphics[width=0.92\columnwidth]{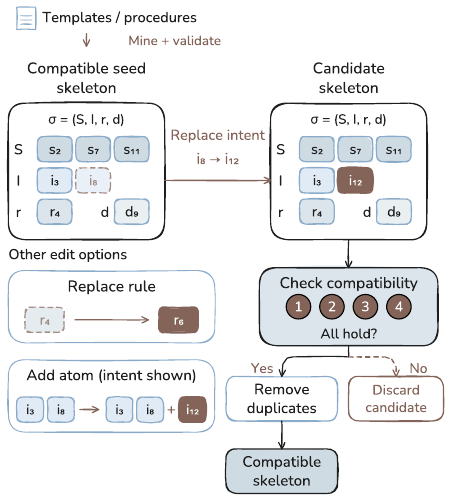}
\caption{Structural recombination. A seed skeleton $\sigma$ mined from a procedure is edited by replacing an intent ($i_8\rightarrow i_{12}$), replacing the rule or adding an atom; the candidate is kept only if constraints 1--4 of Figure~\ref{fig:atoms} all hold, and duplicates are removed.}
\label{fig:recombine}
\end{figure}

The collected atoms of the four families form the \emph{atom pool}, and the closed sets of names they draw on, the operation kinds, verifier kinds, archetypes and carriers of \S\ref{sec:atoms}, form the \emph{vocabulary} $\mathcal{V}$. A skeleton $\sigma=(S,I,r,d)$ picks from the pool a candidate workflow (Figure~\ref{fig:recombine}): up to six servers $S$, the intent atoms $I$, a rule atom $r$ and a dataset $d$. Its parts must satisfy four structural constraints:
\begin{itemize}\itemsep1pt
\item \textbf{Tool coverage.} Every intent's operation kind is implemented by a tool of its server in $S$.
\item \textbf{Carrier hosting.} Some server in $S$ hosts the carrier of $r$.
\item \textbf{Schema compatibility.} $d$ supports the archetype of $r$.
\item \textbf{Rule attachment.} Some intent in $I$ carries $r$.
\end{itemize}
A validator enforces tool coverage and carrier hosting on the skeleton. Schema compatibility is checked against the dataset card while the environment is built, and rule attachment is a requirement placed on the builder (\S\ref{sec:admission}); Appendix~\ref{app:agents} measures how often random compositions satisfy each.

\paragraph{Seed mining.} These constraints make uniform sampling from the atom pool inefficient: of 10{,}000 skeletons assembled at random from the pool, with kinds drawn from those each server implements, 13--14\% satisfy all four constraints (1.6--1.7\% if every intent must carry the rule), and with the rule atom removed all 10{,}000 pass (Appendix~\ref{app:agents}). We therefore mine seed skeletons from real workflow descriptions. Given a business domain $D$ and the atom vocabulary $\mathcal{V}$, which fixes the valid operation kinds, verifier kinds, archetypes and carriers, a mining agent reads public automation templates and operating procedures (Appendix~\ref{app:sources}), maps each step onto $\mathcal{V}$, and writes one skeleton per workflow together with the page it came from:
\[
\mathrm{Mine}(D,\mathcal{V}) \longrightarrow \sigma_{\mathrm{seed}}=(S,I,r,d).
\]
The validator checks each $\sigma_{\mathrm{seed}}$ and returns the violated constraint to the agent, which revises the skeleton until it is accepted. This produced 1{,}394 seed skeletons.

\paragraph{Recombination.} Seeds are then edited within their domain: an intent atom is replaced, the rule atom is replaced, or an atom is added, with the rest of the skeleton held fixed; the result must again pass the validator, and duplicates are removed. Recombination produced 3{,}400 further skeletons.

\begin{table*}[t]
\centering
\scriptsize
\setlength{\tabcolsep}{2.4pt}
\begin{tabular}{l rrrrrr r rrrrr r rrrrrrr r}
\toprule
& \multicolumn{7}{c}{\bench} & \multicolumn{6}{c}{MCP-Mark} & \multicolumn{8}{c}{BFCL v4} \\
\cmidrule(lr){2-8}\cmidrule(lr){9-14}\cmidrule(lr){15-22}
Model & Fin. & HR & Mkt. & Ops. & Sal. & Sup. & \textbf{Avg}
      & File & GitH. & Not. & Play. & Post. & \textbf{Avg}
      & Srch. & Mem. & MultT. & NonL. & Live & Rel. & Irr. & \textbf{Avg} \\
\midrule
\rowcolor{feInk!8}\multicolumn{22}{l}{\emph{Open-weight}}\\
DeepSeek-V4.1-Flash & 54.0 & 58.0 & 57.0 & 60.0 & 46.0 & 32.0 & 51.2
                  & 61.7 & 48.9 & 79.5 & 57.0 & 84.5 & 66.1
                  & 54.0 & 79.8 & 51.5 & 86.4 & 78.4 & 93.8 & 64.7 & 65.2 \\
GLM-5.3-Flash     & 49.0 & 48.0 & 54.0 & 55.0 & 45.0 & 34.0 & 47.5
                  & 82.5 & 83.7 & 88.4 & 74.0 & 89.3 & 83.5
                  & 79.5 & 78.9 & 55.9 & 85.8 & 79.9 & 79.7 & 71.4 & 72.2 \\
DeepSeek-V4-Pro\textsuperscript{$\diamond$} & 16.0 & 22.0 & 36.0 & 30.0 & 25.0 & 29.0 & 26.3
                  & 76.7 & 45.7 & 72.3 & 71.0 & 92.9 & 71.7
                  & 72.0 & 80.4 & 53.4 & 85.0 & 77.2 & 85.9 & 70.1 & 69.7 \\
DeepSeek-V4-Flash\textsuperscript{$\ddagger$} & 30.0 & 32.0 & 35.0 & 47.0 & 23.0 & 27.0 & 32.3
                  & 75.8 & 68.5 & 56.3 & 42.0 & 86.9 & 65.4
                  & 76.0 & 72.5 & 52.6 & 87.0 & 78.1 & 89.1 & 43.9 & 66.4 \\
GLM-5.2\textsuperscript{$\diamond$} & 28.0 & 30.0 & 27.0 & 34.0 & 16.0 & 19.0 & 25.7
                  & 75.0 & 75.0 & 44.6 & 76.0 & 84.5 & 70.1
                  & 71.5 & 66.2 & 56.9 & 88.5 & 79.4 & 92.2 & 76.4 & 69.0 \\
Kimi-K2.6\textsuperscript{$\diamond$} & 18.0 & 12.0 & 15.0 & 24.0 & 12.0 & 14.0 & 15.8
                  & 69.2 & 62.0 & 72.8 & 76.0 & 85.7 & 72.7
                  & 65.0 & 55.7 & 62.1 & 85.2 & 78.1 & 62.5 & 88.4 & 67.9 \\
HY-3              & 8.0 & 10.0 & 12.0 & 17.0 & 16.0 & 16.0 & 13.2
                  & 70.8 & 71.7 & 71.4 & 65.0 & 90.5 & 73.2
                  & 60.5 & 71.2 & 56.6 & 84.9 & 78.7 & 85.9 & 70.2 & 66.7 \\
Qwen3.5-379B      & 7.0 & 5.0 & 2.0 & 10.0 & 7.0 & 5.0 & 6.0
                  & 24.2 & 51.1 & 37.5 & 64.0 & 64.3 & 46.5
                  & 61.5 & 60.9 & 64.0 & 89.5 & 83.1 & 87.5 & 85.4 & 69.5 \\
\midrule
\rowcolor{feInk!8}\multicolumn{22}{l}{\emph{Ours}}\\
\rowcolor{feBlue!14}\basea\textsuperscript{$\S$}                  & 4.0 & 3.0 & 1.0 & 4.0 & 4.0 & 3.0 & 3.2
                        & 45.0 & 40.2 & 33.0 & 55.0 & 53.6 & 44.9
                        & 59.3 & 56.6 & 63.3 & 89.1 & 82.5 & 87.5 & 83.2 & 67.6 \\
\rowcolor{feBlue!32}\quad $+$ \method (SFT) & 23.0 & 24.0 & 26.0 & 32.0 & 19.0 & 15.0 & \textbf{23.2}
                        & 61.7 & 62.0 & 67.9 & 46.0 & 61.9 & \textbf{60.0}
                        & 65.0 & 72.3 & 54.1 & 84.6 & 79.2 & 90.6 & 75.0 & 67.6 \\
\addlinespace
\rowcolor{feBrown!10}\baseb\textsuperscript{$\S$}                  & 19.0 & 17.0 & 10.0 & 21.0 & 11.0 & 18.0 & 16.0
                        & 40.0 & 45.7 & 43.8 & 20.0 & 67.9 & 42.5
                        & 27.9 & 61.0 & 65.0 & 89.0 & 83.8 & 75.0 & 84.6 & 63.0 \\
\rowcolor{feBrown!24}\quad $+$ \method (SFT) & 31.5 & 34.5 & 31.0 & 39.0 & 31.0 & 25.0 & \textbf{32.0}
                        & 65.8 & 65.2 & 71.4 & 64.0 & 75.0 & \textbf{68.5}
                        & 67.5 & 73.8 & 65.1 & 86.4 & 79.9 & 81.3 & 79.4 & \textbf{72.4} \\
\addlinespace
\rowcolor{feGrey!14}\basec\textsuperscript{$\dagger$}                  & 42.5 & 44.0 & 32.5 & 45.0 & 34.0 & 27.5 & 37.6
                        & 63.3 & 71.7 & 80.4 & 68.0 & 85.7 & 73.2
                        & 65.4 & 78.2 & 59.1 & 86.0 & 79.9 & 76.6 & 79.7 & 71.0 \\
\rowcolor{feGrey!32}\quad $+$ \method (RL)  & 45.5 & 47.5 & 37.0 & 45.5 & 36.0 & 30.0 & \textbf{40.3}
                        & 71.7 & 75.0 & 83.0 & 74.0 & 85.7 & \textbf{77.6}
                        & 67.4 & 79.6 & 59.8 & 86.3 & 80.6 & 84.4 & 79.4 & \textbf{72.0} \\
\bottomrule
\end{tabular}
\caption{Main results. All numbers are measured by us: \bench harness 1.0.6, \textsc{api} toolset, 50 steps, one rollout per task, 600 tasks (6 domains $\times$ 100); MCP-Mark, 127 tasks over five MCP servers; BFCL v4, all seven official categories plus the official weighted average. Our rows' \bench cells are means of one to four evaluations (Table~\ref{tab:replay}), the baselines' single evaluations. \textsuperscript{$\diamond$}Served through a commercial gateway that drops the model's reasoning between tool calls, which lowers some models by several points (Appendix~\ref{app:expdetails}); not comparable with the other rows. MCP-Mark and BFCL v4 cells are means of four evaluations per model, whose standard deviations are in Table~\ref{tab:std}. \emph{Rel.} is computed over only 16 tasks and should be read as noise. Open-weight baselines are run at their highest available reasoning setting. \textsuperscript{$\dagger$}teacher that collected the SFT trajectories and the base for RL; \textsuperscript{$\S$}students; \textsuperscript{$\ddagger$}the model that ran every stage of the pipeline, including building the environments. Each \method row is one checkpoint evaluated on all three suites: \basea trained on the round-1$+$3k corpus (5{,}229 trajectories), \baseb on the round-2 corpus (6{,}110), \basec at GRPO step 200 (\S\ref{sec:corpora}, Appendix~\ref{app:expdetails}). The round-3 corpus (34.8\%) scores higher on \bench but lower on MCP-Mark and BFCL v4; GRPO step 200 (40.3\%) scores 1.9 above step 240 on \bench and 1.9 and 1.6 below it on MCP-Mark and BFCL v4 (Appendix~\ref{app:expdetails}).}
\label{tab:main}
\end{table*}

\subsection{Building and admission}
\label{sec:admission}

A builder agent (brief in Appendix~\ref{app:prompts}) turns $\sigma$ into an environment $E=(W_0,q,V,\pi)$: an initial world $W_0$ seeded from $d$, with the rule's text in its carrier and distractor records that satisfy the rule in all but one clause; a request $q$; a verifier $V=V^+\cup V^-$, positive assertions for what must be written and negative ones for what must be left untouched; and a reference solution $\pi$, a sequence of tool calls. The builder runs its program, reads the errors and revises until $E$ passes the \emph{verifier correctness check} below and is admitted.

Write $W_\pi=\pi(W_0)$ for the world after the reference solution and $\pi\setminus s$ for $\pi$ with one write step $s$ removed. $E$ is admitted when
\begin{align}
&\forall a\in V^+\!:\ \neg a(W_0),\quad \forall a\in V^-\!:\ a(W_0); \label{eq:seed}\\
&\forall a\in V:\ a(W_\pi); \label{eq:final}\\
&\forall a\in V^+\ \ \exists s\in\pi:\ \neg a\big((\pi\setminus s)(W_0)\big); \label{eq:witness}\\
&\forall a\in V^-\ \ \exists W':\ \neg a(W'); \label{eq:vacuous}\\
&\mathrm{reset}(W_\pi)=W_0. \label{eq:reset}
\end{align}
Conditions \eqref{eq:seed} and \eqref{eq:final} exclude a verifier that is already satisfied or that nothing can satisfy; \eqref{eq:witness} and \eqref{eq:vacuous} exclude assertions that test nothing; \eqref{eq:reset} makes every attempt start from the same world. An admitted $E$ makes a later failure informative: if $\pi$ passes and a capable agent does not, the fault is in the task or in the agent, and $V$ says which, unless $V$ itself over-specifies the request (\S\ref{sec:repair}).

\subsection{Rollouts and self-evolution}
\label{sec:evolve}

To test whether an admitted environment can be solved, a teacher agent attempts it $N=8$ times. Trajectories that $V$ accepts enter the supervised corpus, and $V$ itself is the reward for reinforcement learning (\S\ref{sec:rl}).

When an environment yields no successful trajectory, a \emph{diagnosis agent} reads $E$, the reference solution $\pi$ and the failed trajectories to find the cause, and one of four pathways follows, the four return arrows of Figure~\ref{fig:overview}:
\begin{itemize}\itemsep1pt
\item \textbf{Repair an atom.} An atom is at fault, for example a server lacks the tool for a required write. An issue is filed against the atom and its patch reaches every later environment built from it.
\item \textbf{Revise the builder.} $V$ demands something that neither $q$ nor $r$ states, or a verifier kind is read two ways. The builder's instructions or the vocabulary are revised, not one environment.
\item \textbf{Supplement the pool.} The failure exposes a missing capability. A server, an operation kind or a dataset is added to the atom pool.
\item \textbf{Keep as a hard task.} The environment is sound and the teacher cannot solve it. It is kept unchanged as a hard task.
\end{itemize}
Through this loop both the atoms and the vocabulary for composing them evolve from round to round (Figure~\ref{fig:scaling}b).

Mining, building, diagnosis, repair and onboarding are all done by agents; Appendix~\ref{app:agents} runs each stage on the model that builds the environments.

\section{Experiments}
\label{sec:exp}

\subsection{Setup}
\label{sec:setup}

\begin{figure*}[t]
\centering
\begin{minipage}[t]{0.37\textwidth}\centering\includegraphics[width=\linewidth]{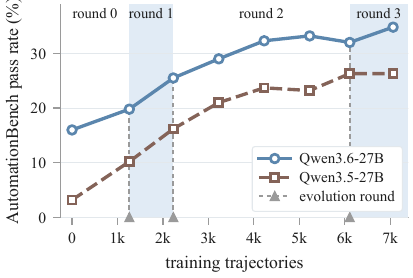}\end{minipage}\hspace{0.02\textwidth}
\begin{minipage}[t]{0.335\textwidth}\centering\includegraphics[width=\linewidth]{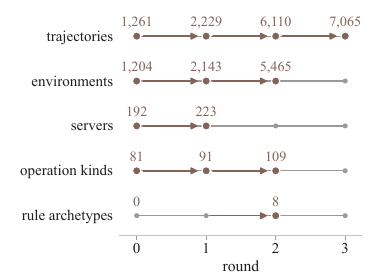}\end{minipage}
\caption{\textbf{(a)} \bench pass rate as the corpus grows. \baseb rises to 34.8\% and \basea to 26.3\%. Bands are evolution rounds. \textbf{(b)} What each round adds; brown marks a value that grew, grey repeats the previous value. Operation kinds are counted under the final taxonomy and rule archetypes are the closed vocabulary introduced in round 2 (Appendix~\ref{app:repairs}).}
\label{fig:scaling}
\end{figure*}

\paragraph{Evaluation.}
We evaluate on \bench~\citep{automationbench}, 600 cross-application business workflows over simulated SaaS applications that share no application with our environments (Appendix~\ref{app:overlap}); MCP-Mark~\citep{mcpmark}, 127 tasks over five MCP servers, none among our atoms; and BFCL v4~\citep{bfcl}, seven function-calling categories including \emph{Irrelevance}, declining a call that should not be made. We run all three under one protocol (Appendix~\ref{app:expdetails}).

\paragraph{Models.}
We train three base models of the same 27B family, a weak, a mid and a strong starting point: their zero-shot scores on \bench are 3.2\%, 16.0\% and 37.6\%, against 51.2\% for the strongest system we measured, DeepSeek-V4.1-Flash.

\basec also serves as the \textbf{teacher}: every supervised trajectory is sampled from it, so the supervised results on \basea and \baseb are distillation from a 37.6\% model, not from a stronger external one.

\paragraph{Corpora.}
\label{sec:corpora}
The corpus grew over the evolution rounds of \S\ref{sec:evolve}, and each round's corpus contains the previous one (Figure~\ref{fig:scaling}b): round 0 covers the environments that can be built on the unrepaired server atoms, round 1 adds those the server repairs made buildable, round 2 adds the environments built from skeletons that carry a rule atom, a family every environment has from this round on, and round 3 adds the repaired environments. Between rounds 1 and 2 we also draw 1k, 2k and 3k of the round-2 additions, stratified by mining batch. The teacher rolls out every environment eight times, a trajectory is kept only if the verifier accepts the final world, and every corpus keeps at most three trajectories per environment, so all points lie on one scale. The round-2 corpus spans 223 applications, 99.5\% of its trajectories cross two or more, and Appendix~\ref{app:corpus} profiles it.

\subsection{Main results}
\label{sec:results}

Table~\ref{tab:main} compares each base with its trained checkpoint on the three suites; every base improves on \bench, the two SFT bases on MCP-Mark, and the two stronger bases on BFCL v4.

\paragraph{Imitation.} The gains on \bench are largest for the two weaker bases: \basea goes from 3.2\% to 23.2\% and \baseb from 16.0\% to 32.0\%. Per task, \basea wins 129 and loses 9 against its base, and \baseb wins 127 and loses 30.

\paragraph{Reward.} \basec is the teacher, so imitation alone is not expected to improve it; the verifiers of the environments, written by DeepSeek-V4-Flash~\citep{deepseekv4} when it built them, can. Used unchanged as a binary reward for GRPO, these verifiers move \basec from 37.6\% to 40.3\% (two evaluations each; 65:53 and 78:58 wins to losses per task), a modest gain on a base that already scores well above DeepSeek-V4-Flash, the model that wrote the verifiers, at 32.3\%.

\paragraph{Transfer.} On MCP-Mark, none of whose servers is among our atoms, \basea gains 15.1 points and \baseb 26.0; on BFCL v4, \baseb gains 9.4 on the official average, most of it on \emph{Search} (27.9 to 67.5), where its base score was an outlier. Two BFCL categories move the other way, its \emph{Irrelevance} and \basea's multi-turn score, and \basec's MCP-Mark gain is within noise (\S\ref{sec:notbuy}).

\subsection{Environments from other pipelines}
\label{sec:baselines}

\begin{table}[t]
\centering
\scriptsize
\setlength{\tabcolsep}{4pt}
\begin{tabular}{@{}lrrrr@{}}
\toprule
Environments & AB & vs.\ ours & MCP-Mark & BFCL v4 \\
\midrule
\method (round 2)             & \textbf{29.5} & --- & \textbf{72.4}$\pm$1.9 & \textbf{68.2}$\pm$0.6 \\
\method (round 1)             & 25.5 & 47:69 & --- & --- \\
AWM~\citep{awm}               & 21.5 & 30:76 & 63.8$\pm$2.3 & 67.6$\pm$0.6 \\
EnvFactory~\citep{envfactory} & 20.4 & 33:81 & 63.8$\pm$2.8 & 66.3$\pm$0.8 \\
EnvScaler~\citep{envscaler}   & 16.8 & 24:97 & 61.4$\pm$1.3 & 63.8$\pm$0.4 \\
EnvCraft~\citep{envcraft}     & 15.8 & 27:100 & 59.1$\pm$1.7 & 67.8$\pm$0.7 \\
\midrule
\baseb, no training           & 16.0 & 27:105 & 42.5$\pm$2.7 & 63.0$\pm$0.4 \\
\bottomrule
\end{tabular}
\caption{One teacher, student, recipe and budget of 2{,}229 trajectories; only the environments differ. \emph{AB}: \bench pass rate, mean of one to four evaluations (Table~\ref{tab:replay}). \emph{vs.\ ours}: per-task wins:losses against the round-2 row, first evaluation of each. \emph{MCP-Mark}, \emph{BFCL v4}: mean $\pm$ s.d.\ of four evaluations (round-1 arm not evaluated). Environment counts, teacher pass rates, turns and token counts per arm are in Table~\ref{tab:arms}.}
\label{tab:baselines}
\end{table}

Table~\ref{tab:baselines} holds the training side fixed and varies the environments across four public pipelines with programmatic verifiers. In each, \basec samples eight attempts per task, the pipeline's own verifier keeps the passing ones (at most three per task), and a random 2{,}229 are drawn; \baseb is trained on each set as in Appendix~\ref{app:expdetails}. Our row is a random 2{,}229 of the round-2 corpus, with the round-1 corpus as a second row. Our round-2 sample scores 29.5\%, 8.0 points above the best of the four (AWM, 21.5\%) and 13.7 above the lowest (EnvCraft, 15.8\%), and the paired counts point the same way on every row: 76:30 against AWM, 81:33 or better against the other three pipelines; on the two transfer suites the same arm leads every pipeline arm by 8.6 points or more on MCP-Mark, about three standard deviations, and by 0.4 to 4.4 on BFCL v4, where the two smallest margins are within noise. The round-1 corpus, 2{,}229 trajectories built before rules became an atom family, scores 25.5\%, so the round that added the rule-atom family is worth 4.0 points at fixed size. The arm drawn from the most environments also scores highest (Table~\ref{tab:arms}), but the arms also differ in turns, tokens and teacher pass rate, and the pipelines in provenance and composition; the comparison bounds the joint effect, not any one atom family.

\section{Analysis}
\label{sec:analysis}

\subsection{Scaling the atom pool}
\label{sec:scaling}

Figure~\ref{fig:scaling}a varies the number of trajectories, and both bases improve as the corpus grows: \baseb rises from 16.0\% to about 35\%, the last four points within three points of one another; \basea rises from 3.2\% to 26.3\%, with one dip of half a point. Figure~\ref{fig:scaling}b shows what each round added.

\subsection{One round of evolution}
\label{sec:repair}

Of the 4{,}662 environments admitted in round 2 (built from 4{,}794 skeletons), 71.6\% yield at least one accepted trajectory in eight teacher attempts; we call this fraction the \emph{yield}. The 1{,}324 that yield nothing look no different from the rest: they passed admission and read as reasonable tasks, so a judge-based pipeline would discard them. Admission (\S\ref{sec:admission}) rejects a vacuous or already-satisfied verifier; one that demands a value the request never states passes, because the reference solution carries it, and only a rollout exposes it.

With an executable verifier and a reference solution the fault can be located, and it is of three kinds. Half (52\%) are verifier faults: the verifier demands something neither the request nor the rule asked for. A quarter (23\%) are world faults: a record, a column or a tool the task needs is missing or inconsistent. The rest (25\%) are sound environments the teacher cannot solve. Repairing the first two kinds and re-sampling lifts the yield from 71.6\% to 88.3\%; the third kind is left unchanged. After re-sampling, 220 of the repaired environments still yield nothing and form the exploration set of \S\ref{sec:rl}. The 997 repairs reduce to 33 root causes (Appendix~\ref{app:rootcauses}), almost all of them a verifier kind read two ways or a builder instruction, so a fix made once reaches every environment that shares it.

\begin{figure}[t]
\centering
\includegraphics[width=\columnwidth]{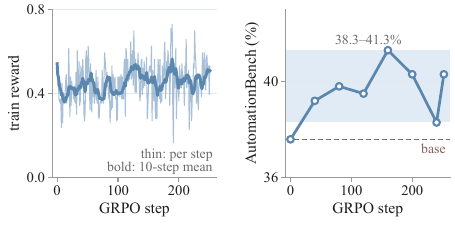}
\caption{GRPO on \basec with the task verifiers as a binary reward. \textbf{Left:} training reward, the fraction of 96 sampled episodes the verifier accepts. \textbf{Right:} \bench pass rate at seven checkpoints against the 37.6\% base; from step 40 on it stays in a 38.3--41.3\% band while reward keeps climbing.}
\label{fig:rl}
\end{figure}

\subsection{Replacing real atoms with generated ones}
\label{sec:ablation}

\begin{table}[t]
\centering\small
\setlength{\tabcolsep}{4pt}
\resizebox{\columnwidth}{!}{%
\begin{tabular}{lcccrr}
\toprule
Atoms & \bench & MCP-Mark & BFCL v4 & env. & tokens \\
\midrule
real & \textbf{30.3}$\pm$0.7 & \textbf{54.9}$\pm$2.7 & \textbf{66.3}$\pm$0.4 & 859 & 7.6M \\
generated data & 28.0$\pm$0.9 & 51.2$\pm$2.8 & 65.2$\pm$0.4 & 759 & 7.3M \\
generated intents & 27.3$\pm$0.5 & 52.4$\pm$2.5 & 65.9$\pm$0.9 & 748 & 7.4M \\
\bottomrule
\end{tabular}
}
\caption{One atom family replaced by model-generated content, the rest fixed; \baseb trained on 885 trajectories per arm. Means $\pm$ standard deviation over four evaluations; one training run per arm. \emph{env.}: distinct environments among the 885 trajectories; \emph{tokens}: supervised tokens.}
\label{tab:ablation}
\end{table}

\S\ref{sec:baselines} varied the whole pipeline. Here we replace one atom family at a time and keep everything else fixed: the builder and its prompt, admission, the teacher and the acceptance rule, the length filter and the SFT recipe. The cohort is the 1{,}326 environments of the round-2 sample that were built from skeletons; the other 812 predate the skeleton of \S\ref{sec:recombine}. \emph{Real atoms} keeps their original trajectories. \emph{Generated data} replaces every table the builder read from a real dataset with one a model writes from the table's schema card (column names, types, at most 5{,}000 rows), without seeing a real row. \emph{Generated intents} replaces the 3{,}471 mined intent atoms with ones a model writes from each server's capability card and the closed vocabulary of operation and verifier kinds, without access to the mined pages; the skeletons are then recombined to the same count with the rule atoms unchanged.

Supply falls first. The builder admits 94.3\% of the generated-data environments and 84.9\% of the generated-intent ones, and the teacher solves 67.0\% and 67.8\% of those at least once; the control is at 100\% by construction. The two arms end with 965 and 885 trajectories, so all three are cut to 885 and trained alike. Real atoms lead on every suite (Table~\ref{tab:ablation}): by 2.2 and 3.0 points on \bench, above its evaluation spread of 0.5--0.9; by 3.7 and 2.6 on MCP-Mark, within its spread of 2.5--2.8; and by 1.1 and 0.4 on BFCL v4, the second within noise. Within BFCL the lead comes from multi-turn use and irrelevance detection; the generated arms do better on memory and web search.

Two differences in content are measured, though neither is isolated as a cause. Generated intents shift the operations the corpus exercises: among the 3{,}471 pool positions, \texttt{create\_record} rises from 51 to 371, \texttt{delete\_or\_archive\_row} from 11 to 332, and \texttt{helpdesk\_reply\_or\_update} falls from 302 to 34. And at equal size both generated arms cover fewer distinct environments, 759 and 748 against 859, because fewer environments yield and more contribute a second trajectory; supervised tokens differ by under 3.5\%. Each arm is one training run, and the MCP-Mark and BFCL v4 scores of this batch were evaluated later than those of Table~\ref{tab:main}, so they are compared only within the table.

\subsection{The same verifiers as a reward}
\label{sec:rl}

Nothing is rebuilt for reinforcement learning: the verifier that admitted an environment is the reward. We run GRPO~\citep{deepseekmath} on the unmodified \basec over 1{,}516 environments: those the teacher solved in one to six of eight attempts, and the 220 it never solved, kept as an exploration set (Appendix~\ref{app:expdetails} profiles both). Replaying the reference solutions earns the reward; a no-op policy does not. The pass rate moves from 37.6\% into a 38.3--41.3\% band by step 40 and oscillates there (Figure~\ref{fig:rl}); every checkpoint scores above the base and, paired against it, wins more tasks than it loses, but the margins are not significant at the 5\% level (Appendix~\ref{app:expdetails}) and no ordering inside the band is established, so we report the band, with Table~\ref{tab:main} giving one point in it. Training reward keeps rising after the pass rate flattens: the policy goes on fitting the training environments after the transferable part has been learned.

\subsection{What the training does not buy}
\label{sec:notbuy}

Three results go the other way or not at all. First, BFCL's \emph{Irrelevance} falls by five to eight points after supervised training on both weaker bases: every request asks for at least one write and the negative assertions forbid particular writes, not the request as a whole, so nothing rewards declining. Second, \basea's BFCL multi-turn score falls from 63.3 to 54.1. Third, the MCP-Mark gain on \basec is not established: 73.2 to 77.6 against standard deviations of 3.7 and 2.6.

\section{Conclusion}
\label{sec:conclusion}
Synthesising an agentic environment is the easy half; the hard half is finding out that it is wrong: 28\% of ours yielded no trajectory, three quarters of those because the environment was at fault. Atoms make such a defect attributable to a reusable unit, and a tested verifier makes a failed rollout a bug report, not a mystery. One round of the loop raised the yield from 71.6\% to 88.3\%; the environments trained three bases, with supervised gains that transfer to MCP-Mark, outscored four public pipelines at equal budget, and as a reward moved a strong base from 37.6\% to 40.3\%.

\section*{Limitations}\label{pg:limits}
Four limits bound what this paper shows.

\paragraph{Scope of the comparison.} The controlled comparison of \S\ref{sec:baselines} covers the four public pipelines that release executable environments with programmatic verifiers; pipelines that release only trajectories, or whose verifier is an LLM judge, cannot be sampled by the same teacher under the same acceptance rule, so they are compared by provenance (Table~\ref{tab:related}, our reading of those papers) rather than trained on. The ablation of \S\ref{sec:ablation} replaces the data atoms or the intent atoms with model-generated ones on the skeleton-built part of the round-2 sample while holding the rest fixed; each arm is one training run, so it indicates rather than isolates the contribution of a family, and the comparison of \S\ref{sec:baselines} still bounds the joint effect of provenance and composition. What comes from real sources is the atoms: the request, the verifier and the reference solution are still written by the builder model, and the diagnosis is made by an agent.

\paragraph{Statistical resolution.} No training run is repeated. \bench scores of the models we serve are means of one to four evaluations of 600 tasks (Table~\ref{tab:replay}), so we report per-task paired counts and a band for reinforcement learning; MCP-Mark and BFCL v4 scores are means of four evaluations whose standard deviations reach 3.7 and 0.8 points (Table~\ref{tab:std}), and we treat differences of that size as unresolved. The paired counts of Table~\ref{tab:baselines} point one way on every row; the closest, 69:47 for round 2 against round 1, is at the 5\% level (two-sided sign test $p\approx0.05$). For reinforcement learning, one of the 18 pairings of a checkpoint evaluation with a base evaluation is significant at that level.

\paragraph{Evaluation protocol.} Three baselines served through a commercial gateway (DeepSeek-V4-Pro, GLM-5.2 and Kimi-K2.6, \textsuperscript{$\diamond$} in Table~\ref{tab:main}) are evaluated with the model's reasoning dropped between tool calls, which lowers some models by several points (5.3 for \basec and 6.5 for DeepSeek-V4-Flash when we measured both ways, Table~\ref{tab:replay}). Every claim of the paper compares rows evaluated under one protocol; the ranking of our rows against those three is not established.

\paragraph{Breadth.} Yields are given for one evolution round, on one 27B family, in English business domains; the loop is not shown to converge.

\label{pg:refs}
\bibliography{custom}

\appendix
\raggedbottom
% Appendix layout: no float-only page unless it is at least 80% full.
\renewcommand{\floatpagefraction}{0.8}
\renewcommand{\dblfloatpagefraction}{0.8}
\section{The condition form}
\label{app:grammar}

A condition is $\langle e,\varphi,\kappa\rangle$. The scope $\varphi$ is a conjunction of clauses $\texttt{col}\ \mathit{op}\ v$ over 14 comparison operators (equality, order, membership, string containment and prefix/suffix, null tests). A value $v$ is a literal (a scalar or a non-scalar constant); a reference into a namespace (\texttt{\$arg}, the call's arguments; \texttt{\$actor}, the caller; \texttt{\$session}; \texttt{\$now}); another column of the same row, or an arithmetic expression over such columns; a sub-query nested at most two deep; or the value a column held before the operation (\texttt{\$was}). The cardinality $\kappa$ is one of $\geq n$, $=n$, $\leq n$, $\neq n$, $>n$, $<n$, with \emph{exists} and \emph{not exists} as aliases. The condition appears at six sites (read filter, entity constraint, precondition, postcondition, provides and graph locator); each fixes which namespaces are visible, a precondition may refer to \texttt{\$arg} and a verifier assertion only to the final world, and how an undecidable comparison is treated. Guards and verifier assertions share one representation, one operator table and one evaluator; entity constraints and read filters use the same representation through adapters. Over the 384 usable servers, 19{,}895 precondition clauses use this form.

\section{Compiling a server atom}
\label{app:compile}

A code-reading agent is given one open-source repository and a contract of about 1{,}200 lines. The contract fixes closed vocabularies for what it may write (7 operation types, 15 effect kinds, 6 entry kinds, 3 edge relations), gives for each a counter-example of what does not belong, and requires absence to be stated: an empty list means ``I checked'', a missing field means ``I did not look''. The agent describes every entity with its columns and constraints, every operation with its parameters and the effects it leaves in the database, every precondition the code enforces and the message the code returns when it is violated, and the relations between operations. A validator scores the description against the contract and returns every violation, with the way to fix it, to the same agent session, and the agent revises. Of the 467 repositories, 433 reached this stage; over four feedback rounds, 395 met the contract; after the second stage, 384 are usable.

A second stage translates the description into an executable specification. Everything that can be translated mechanically is, and every default the translator had to guess is listed for the agent to confirm; what requires reading code again, such as which entity is the account table and which guards are real, is decided by the agent. The runtime serves each specification as a set of tools over a relational state, dispatching by operation family, and evaluates every precondition through the condition form of Appendix~\ref{app:grammar}. The usable set holds 384 servers with 25{,}689 tools; of the 14{,}810 tools with a precondition, one lacks a refusal message.

\section{Atom schemas}
\label{app:atoms}

\paragraph{Intent atom.} \texttt{\{app, kind, pos, neg, check\_kinds\}}: the server (\texttt{app} in the skeleton schema, \texttt{service} in the compiler contract); the operation kind, which must be implemented by a write tool of that server; the number of positive and of negative predicates the verifier should carry for this atom (1--30 and 0--30); and the verifier kinds a verifier may use. A real instance:
\begin{schemabox}{INTENT ATOM --- REAL INSTANCE}
\{"app": "espocrm", "kind": "attach\_note\_or\_task", "pos": 4, "neg": 3, "check\_kinds": ["note\_or\_task\_exists", "note\_or\_task\_not\_exists"]\}
\end{schemabox}

\paragraph{Rule atom.} \texttt{\{archetype, carrier, has\_exception, rule\_shape\}}. The shape is at most 400 characters and may contain no proper names and no number of two or more digits; the validator rejects the numbers, the builder instructions forbid the names. A real instance:
\begin{schemabox}{RULE ATOM --- REAL INSTANCE}
\{"archetype": "approval\_chain", "carrier": "notes\_field", "has\_exception": true, "rule\_shape": "Record-correction requests with disputed factual fields enter amendment review; denied cases receive a hearing-option task instead of silent closure."\}
\end{schemabox}
Carriers and the servers registered to host them: policy sheet (budibase, directus, ledgersmb), policy document (outline, paperless-ngx), policy mail (stalwart), note field (14 servers, among them CRM, ticketing, accounting and booking applications), card description (vikunja, mantisbt, planka), pinned channel message (zulip).

\paragraph{Data atom.} A dataset directory with a card: for each table with at least eight rows, the type of every column (numeric, date, category, text, identifier, contact, entity), its number of distinct values and sample values; the rule archetypes the table can support; size and licence.

\paragraph{Server atom.} The executable specification: entities with columns and constraints, tools with parameters, effects and preconditions in the condition form, an account entity with a known credential, and, once onboarded, a capability card mapping operation kinds to the tools that implement them.

\paragraph{Skeleton.} \texttt{\{domain, apps, deliverable\_atoms, rule\_atoms, dataset\_hint, n\_intents, n\_distractors\_min, predicate\_budget, chain, source\}}: the business domain; one to six servers (one to five in the corpus); the intent atoms (\texttt{deliverable\_atoms} in the code) and one to three rule atoms (one or two in the corpus); the dataset, left empty when the builder is to choose one from the pool, with schema compatibility checked then (\S\ref{sec:recombine}); the number of intents, the minimum number of distractor records, the number of predicates (8--60) the verifier must at least reach, which 95.3\% of skeletons do; a one-line chain from reading the carrier to the last write; and the public URL the skeleton was mined from or the skeleton it was recombined from. The skeleton that the two atoms above belong to, mined from a public page on student-record rights (bookkeeping fields omitted):
\begin{schemabox}{SKELETON --- REAL INSTANCE}
\{"skeleton\_id": "c\_mine\_education\_008", "domain": "education", "source": "web:https://studentprivacy.ed.gov/ferpa", "apps": ["espocrm", "gladys"], "deliverable\_atoms": [\{"app": "espocrm", "kind": "attach\_note\_or\_task", "pos": 4, "neg": 3, "check\_kinds": ["note\_or\_task\_exists", "note\_or\_task\_not\_exists"]\}, \{"app": "gladys", "kind": "schedule\_event", "pos": 3, "neg": 2, "check\_kinds": ["event\_exists\_with", "event\_not\_exists"]\}], "rule\_atoms": [\{"archetype": "approval\_chain", "carrier": "notes\_field", "has\_exception": true, "rule\_shape": "Record-correction requests with disputed factual fields enter amendment review; denied cases receive a hearing-option task instead of silent closure."\}], "n\_intents": 2, "n\_distractors\_min": 2, "predicate\_budget": 12, "delivery\_shape": "per-record actions and auditable follow-up", "chain": ["Read the notes field to learn the applicable rule.", "Compare the source records and skip look-alike exceptions.", "Apply each listed business write to eligible records.", "Leave a concise record of the selected and excluded cases."], "dataset\_hint": []\}
\end{schemabox}

\section{Inducing the verifier kinds}
\label{app:verifierkinds}

To test whether the 45 verifier kinds can be recovered from public documentation of real work, an agent on the model that builds the environments was run over 115 public pages of business-workflow documentation (vendor API references, help centres and procedure documents), several sessions per page. Each session received the condition form of \S\ref{sec:grammar}, the entity and tool metadata of the servers, and one starting page chosen by a coordinator; for nine of the 45 kinds the coordinator also pre-extracted public field names or schema facts from those pages. No session received the 45 kinds or another session's proposals; a scan of all 5{,}609 input files confirms this. Each session wrote down, for the workflow it read, what would have to hold in the final world, as a named family with a template over the condition form, a quotation from the source and a positive and a negative example.

\begin{figure*}[t]
\centering
\includegraphics[width=\textwidth]{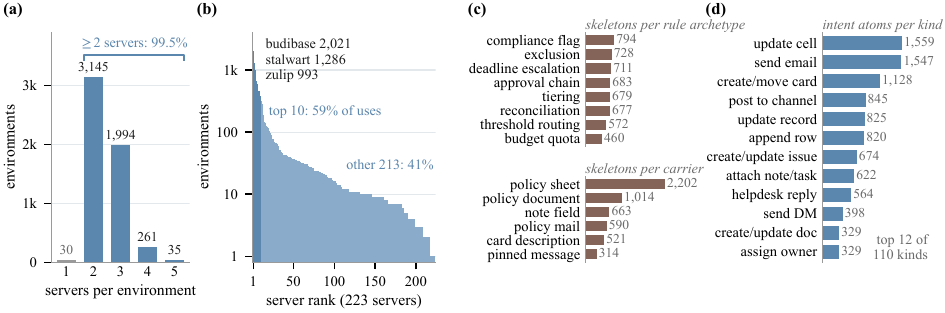}
\caption{The corpus. (a) Servers per environment; almost every environment spans two or more. (b) Use of each server, ranked; a dozen servers for spreadsheets, mail, chat and boards carry most of the load, as they do in real automation templates. (c) Rule archetypes and carriers over the 4{,}794 skeletons. (d) The twelve most frequent of the 110 operation kinds.}
\label{fig:corpus}
\end{figure*}

The sessions produced 1{,}881 family proposals, 405 of them complete. Judged against the 45 verifier kinds, the proposals are semantically equivalent to all 45; under a strict reading that requires the same generic template rather than a specialisation, 3 of 45 were confirmed in the 14 decisions reviewed, so the strict figure is a lower bound and not an estimate. Of 64 quotations sampled across the 45 kinds, 64 were found verbatim in the cached source page. A coordinator who knew the target vocabulary chose the starting pages, so the result shows that the vocabulary is recoverable from public sources given where to look, not that it would be discovered without guidance.

\section{Experimental details}
\label{app:expdetails}

\begin{table*}[t]
\centering
\footnotesize
\setlength{\tabcolsep}{3.5pt}
\begin{tabular}{@{}llrlrl@{}}
\toprule
Base & Checkpoint & Dropped & Replayed & $n$ & vs.\ base \\
\midrule
\basea & base & 2.7 & 3.2 & 1 & \\
 & round 0 (1{,}261) & 5.7 & 10.2 & 1 & \\
 & round 1 (2{,}229) & 9.0 & 16.2 & 1 & \\
 & 3{,}229 & 16.0 & 21.0 & 1 & \\
 & 4{,}228 & 16.8 & 23.7 & 1 & \\
 & round 1$+$3k (5{,}229)$^\star$ & 21.0 & 23.2 & 1 & 129:9 \\
 & round 2 (6{,}110) & 17.7 & 26.3 & 1 & 145:6 \\
 & round 3 (7{,}065) & 18.3 & 26.3 & 1 & 146:7 \\
\midrule
\baseb & base & 14.2 & 16.0 & 1 & \\
 & round 0 (1{,}261) & 19.0 & 19.8 & 1 & \\
 & round 1 (2{,}229) & 23.7 & 25.5$\pm$0.2 & 2 & \\
 & 3{,}229 & 29.8 & 29.0 & 1 & \\
 & 4{,}228 & 29.2 & 32.3 & 1 & \\
 & round 1$+$3k (5{,}229) & 27.0 & 33.2 & 1 & 133:30 \\
 & round 2 (6{,}110)$^\star$ & 29.2 & 32.0$\pm$0.2 & 2 & 127:30 \\
 & round 3 (7{,}065) & 30.2 & 34.8 & 1 & 132:19 \\
\addlinespace
 & Table~\ref{tab:baselines}: ours (round 2) & 26.5 & 29.5$\pm$1.0 & 4 & \\
 & Table~\ref{tab:baselines}: round 1 & 23.7 & 25.5$\pm$0.2 & 2 & 69:47 \\
 & Table~\ref{tab:baselines}: AWM & 23.2 & 21.5$\pm$0.2 & 4 & 76:30 \\
 & Table~\ref{tab:baselines}: EnvFactory & 20.3 & 20.4$\pm$0.9 & 3 & 81:33 \\
 & Table~\ref{tab:baselines}: EnvScaler & 18.3 & 16.8$\pm$0.0 & 2 & 97:24 \\
 & Table~\ref{tab:baselines}: EnvCraft & 16.3 & 15.8$\pm$1.4 & 2 & 100:27 \\
\midrule
\basec & base & 32.3 & 37.6$\pm$0.8 & 2 & \\
 & GRPO step 40 & 34.0 & 39.2 & 1 & 66:60 \\
 & GRPO step 80 & 38.2 & 39.8 & 1 & 63:53 \\
 & GRPO step 120 & 35.5 & 39.5 & 1 & 66:58 \\
 & GRPO step 160 & 37.2 & 41.3 & 1 & 71:52 \\
 & GRPO step 200$^\star$ & 38.5 & 40.3$\pm$0.1 & 2 & 65:53 \\
 & GRPO step 240 & 38.0 & 38.3$\pm$2.1 & 2 & 56:64 \\
 & GRPO step 252 & 35.8 & 40.3 & 1 & 62:49 \\
\midrule
DeepSeek-V4-Flash & & 25.8 & 32.3 & 1 & \\
\bottomrule
\end{tabular}
\caption{\bench pass rate with the model's reasoning dropped between tool calls (an earlier configuration of our gateway, one evaluation each) and with it replayed (the protocol of Tables~\ref{tab:main} and \ref{tab:baselines}); $n$ is the number of replayed evaluations, with mean and standard deviation where $n>1$. \emph{vs.\ base}: tasks won and lost against the base under the replayed protocol, first evaluation of each; in the Table~\ref{tab:baselines} block, tasks won and lost by our round-2 arm against that row. Checkpoints are named by their corpus (\S\ref{sec:corpora}), with trajectory counts in parentheses; $^\star$ marks the rows of Table~\ref{tab:main}.}
\label{tab:replay}
\end{table*}

\paragraph{Harness.} We run the unmodified upstream harness at version 1.0.6 on the public split of 600 scored tasks (six domains $\times$ 100: Sales, Marketing, Operations, Support, Finance, HR). Four settings matter for comparability; we fix the hardest of the first three. (i) \emph{Toolset}: the \textsc{api} toolset, which exposes only \texttt{api\_search}, \texttt{api\_fetch} and \texttt{base64\_encode}, so the agent must discover endpoints and construct requests itself; the benchmark's authors report that its higher-level toolsets usually score higher. (ii) \emph{Step budget}: the default cap of 50 agent steps; across the audited runs 15.9\% of episodes reach it. (iii) \emph{Coverage}: a single rollout per task, with no best-of-$n$ or majority vote. (iv) \emph{Reasoning}: the models we serve ourselves (every \method row, its base, and the baselines DeepSeek-V4.1-Flash, GLM-5.3-Flash, DeepSeek-V4-Flash, HY-3 and Qwen3.5-379B) run from an inference endpoint with thinking enabled, and the unmodified harness keeps the model's reasoning of earlier turns in the conversation; DeepSeek-V4-Pro, GLM-5.2 and Kimi-K2.6 are served through a commercial gateway that rejects the reasoning of earlier turns in the request, so they receive the conversation without it. An earlier configuration of our own gateway dropped it too, and the paragraph \emph{Reasoning replay} below reports both measurements for the models we serve. We audited the graded tasks behind 24 runs under the earlier configuration and the 45 runs behind Table~\ref{tab:replay}: every pass satisfies \texttt{assertions\_passed == assertions\_total}, and no episode was aborted by the inference gateway or left with an unresolved request fault.

\paragraph{Suites.} \bench poses six domains $\times$ 100 tasks over simulated SaaS applications whose REST endpoints the agent must discover, with layered business rules and deliberately misleading records; we use the public split, the \textsc{api} toolset, a cap of 50 steps per task and one rollout, and report the official \texttt{task\_completed\_correctly}, whose mean over tasks is the pass rate. MCP-Mark's five servers are filesystem, GitHub, Notion, Playwright and PostgreSQL, each task with a curated initial state and a validation script. For BFCL v4 we report all seven official categories (including multi-turn use and long-context memory) and the official weighted average.

\paragraph{Checkpoint selection.} Each \method row of Table~\ref{tab:main} is one checkpoint evaluated on all three suites. \baseb reaches 34.8\% on \bench with the round-3 corpus but scores lower on MCP-Mark and BFCL v4 (59.1\% and 67.9, single evaluations) than the round-2 checkpoint in the table (68.5\% and 72.4), so the table reports the latter. For \basea, the round-2 and round-3 corpora both score 26.3\% against 23.2\% for the round-1$+$3k checkpoint in the table; the round-3 checkpoint matches it on MCP-Mark and BFCL v4 (60.6\% and 67.6 over three evaluations against 60.0\% and 67.6 over four), so the ordering is not established and the table reports the checkpoint evaluated four times. For \basec, GRPO steps 200 and 240 score 40.3\% and 38.3\% on \bench (two evaluations each; 65:53 and 56:64 per task against the first base evaluation), 77.6\% and 79.5\% on MCP-Mark and 72.0 and 73.6 on BFCL v4, step 240 evaluated once there; the MCP-Mark difference is inside the evaluation spread, the BFCL v4 difference of 1.6 favours step 240, and no ordering inside the band is established (\S\ref{sec:rl}); the table reports step 200, the checkpoint evaluated four times on MCP-Mark and BFCL v4.

\paragraph{Evaluation spread.} Every model of Table~\ref{tab:main} was evaluated four times on MCP-Mark and on BFCL v4 under the same configuration, and the table reports the means; Table~\ref{tab:std} gives the standard deviation of the MCP-Mark pass rate and of the BFCL v4 weighted average. The number of \bench evaluations behind each checkpoint is listed in Table~\ref{tab:replay}.

\paragraph{Reasoning replay.} Served from an inference endpoint with thinking enabled, the unmodified harness keeps the model's reasoning of earlier turns in the conversation; an earlier configuration of our gateway dropped it, as the commercial gateway behind the \textsuperscript{$\diamond$} rows of Table~\ref{tab:main} does. Table~\ref{tab:replay} reports both for every checkpoint we serve: the \emph{dropped} column is one evaluation each under the earlier configuration, the \emph{replayed} column the evaluations behind Tables~\ref{tab:main} and \ref{tab:baselines}. The bases are the most sensitive rows: \basec gains 5.3 points from the replayed reasoning and DeepSeek-V4-Flash 6.5, the two smaller bases 0.5 and 1.8. The checkpoints trained on our environments move by $-0.8$ to $+8.6$ points and the four pipeline arms by $-1.7$ to $+0.1$, so every supervised gain and the ordering of Table~\ref{tab:baselines} hold under both protocols, with the lead over the closest pipeline at 3.3 points dropped and 8.0 replayed. The GRPO checkpoints, trained with their reasoning kept, gain 0.3 to 5.2 (3.1 on average), less than their base, so the reinforcement-learning gain is 6.2 points dropped and 2.7 replayed; of the 18 pairings of a checkpoint evaluation with a base evaluation, 16 win more tasks than they lose, one at the 5\% level (step 160 against the second base evaluation, 77:51, $p\approx0.03$). Scores in the two columns are not comparable with each other, and neither is comparable with published numbers taken under a third configuration.

\begin{table}[t]
\centering
\footnotesize
\setlength{\tabcolsep}{4pt}
\begin{tabular}{@{}lrr@{}}
\toprule
Model & MCP-Mark & BFCL v4 \\
\midrule
DeepSeek-V4.1-Flash & 0.6 & 0.6 \\
GLM-5.3-Flash & 3.4 & 0.4 \\
DeepSeek-V4-Pro & 1.9 & 0.7 \\
DeepSeek-V4-Flash & 2.3 & 0.4 \\
GLM-5.2 & 2.6 & 0.5 \\
Kimi-K2.6 & 2.0 & 0.8 \\
HY-3 & 2.8 & 0.7 \\
Qwen3.5-379B & 2.2 & 0.6 \\
\midrule
\basea & 1.7 & 0.5 \\
\quad $+$ \method (SFT) & 2.0 & 0.3 \\
\baseb & 2.7 & 0.4 \\
\quad $+$ \method (SFT) & 2.9 & 0.5 \\
\basec & 3.7 & 0.7 \\
\quad $+$ \method (RL) & 2.6 & 0.3 \\
\bottomrule
\end{tabular}
\caption{Standard deviation, in points, of the MCP-Mark pass rate and the BFCL v4 weighted average over the four evaluations behind each row of Table~\ref{tab:main}.}
\label{tab:std}
\end{table}

\paragraph{Baselines.} Every baseline in Table~\ref{tab:main} is an open-weight model run by us under the harness above, at its highest available reasoning setting, five from our own endpoints and three (\textsuperscript{$\diamond$}) through a commercial inference gateway; we quote no leaderboard numbers. Our GLM-5.2 score of 25.7\% against the 26.2\% on the benchmark's public leaderboard is a check that the configuration reproduces the authors'.

\subsubsection*{Training}
\paragraph{SFT.} 64 H20 GPUs, 3 epochs, global batch size 32, peak learning rate $1\times10^{-5}$ with cosine decay, 64K sequence length. We export a checkpoint at each epoch boundary and evaluate the last unless stated otherwise. Assistant turns that receive a 4xx tool response are excluded from the loss.

\begin{figure}[t]
\centering
\includegraphics[width=0.8\columnwidth]{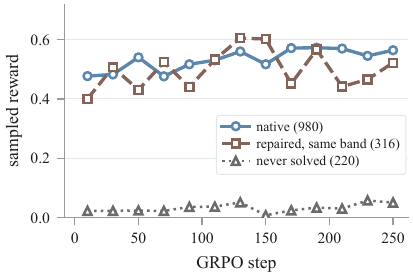}
\caption{Training reward by prompt group during GRPO, in bins of 20 steps: the 980 native environments, the 316 repaired environments in the same teacher band, and the 220 of the exploration set.}
\label{fig:rlgroups}
\end{figure}

\paragraph{RL.} We run GRPO~\citep{deepseekmath} on \basec, initialised from the \emph{unmodified} base rather than from an SFT checkpoint, so the RL result is not confounded with the supervised one. 128 H20 GPUs (64 training, 64 rollout), group size 8, temperature 1.0, 2 epochs over 1{,}516 prompts, giving 252 optimisation steps ($\approx$34 h). The reward is exactly the task's programmatic verifier: binary, unshaped, with no reward model, no LLM judge and no auxiliary terms. \S\ref{sec:rl} describes the prompt set. Of the 1{,}516 prompts, 536 (35.4\%) exist only because of the round-3 repairs, among them all 220 of the exploration set. In the same teacher band (one to six successes in eight) the repaired environments train like the native ones: over the run the sampled reward rises from 0.40 to 0.52 for the 316 repaired environments and from 0.48 to 0.56 for the 980 native ones, and the two curves cross throughout (Figure~\ref{fig:rlgroups}). The exploration set is sampled like every other prompt (eight rollouts per epoch, sixteen over the run): 48 of the 220 are solved at least once during training, the first success at a median of step 91 of 252, and 172 remain unsolved at the end.

\paragraph{Pipeline comparison.} EnvFactory tasks are single turns of released sessions, with the session prefix given as context and excluded from the loss (median 9 assistant turns including the prefix); AWM trajectories are filtered by its code-only verifier; the LLM judge of its released pipeline is not used, and AWM's authors report the code-only option as the weaker of the two (Appendix~\ref{app:related}). All arms of Table~\ref{tab:baselines} share the base checkpoint, learning rate, epochs, sequence length and tool-call template; the training configurations differ only in the data path. Trajectories are encoded with the teacher's reasoning preserved across turns. Table~\ref{tab:arms} profiles five of them; EnvFactory's teacher pass@8 is 0.72 over all rows and 0.59 over the rows kept. All four pipelines are served through the same tool interface as our environments; the mapping is structural and changes neither tool definitions nor content. Per domain, our arm leads AWM in all six domains (mean of four evaluations each). On the two transfer suites our arm again ranks first: its student scores 72.4\% on MCP-Mark and 68.2 on BFCL v4, against 59.1--63.8\% and 63.8--67.8 for the four other arms, means of four evaluations (Table~\ref{tab:baselines}).

\begin{table}[t]
\centering
\footnotesize
\setlength{\tabcolsep}{4pt}
\resizebox{\columnwidth}{!}{%
\begin{tabular}{@{}lrrrrrrr@{}}
\toprule
Arm & pass@8 & env. & turns & tokens & loss & MCP & BFCL \\
\midrule
\method (round 2) & --- & 1{,}173 & 13 & 43.3M & 14.3M & \textbf{72.4} & \textbf{68.2} \\
AWM        & 0.59 & 556 & 6  & 40.4M & 7.6M & 63.8 & 67.6 \\
EnvFactory & 0.72 & 142 & 4  & 21.4M & 5.5M & 63.8 & 66.3 \\
EnvScaler  & 0.36 & 47  & 8  & 41.1M & 13.9M & 61.4 & 63.8 \\
EnvCraft   & 0.87 & 61  & 18 & 22.5M & 9.1M & 59.1 & 67.8 \\
\midrule
\baseb (no training) & --- & --- & --- & --- & --- & 42.5 & 63.0 \\
\bottomrule
\end{tabular}}
\caption{Five of the arms of Table~\ref{tab:baselines}, 2{,}229 trajectories each, over 2{,}138, 767, 818, 763 and 768 distinct tasks respectively. \emph{pass@8}: share of tasks the teacher solved at least once in eight attempts. \emph{env.}: distinct worlds (\texttt{env\_id}) behind the trajectories; several of our tasks share a world. \emph{turns}: median assistant turns that carry loss (EnvFactory's prefix turns, median 9 including them, carry none). \emph{tokens}, \emph{loss}: total tokens and tokens that carry loss. \emph{MCP}, \emph{BFCL}: MCP-Mark pass rate and BFCL v4 weighted average of the trained student, means of four evaluations as in Table~\ref{tab:baselines}.}
\label{tab:arms}
\end{table}

\section{Overlap with \bench}
\label{app:overlap}

Of the transfer suites, BFCL v4 touches the corpus at one point: eight environments (0.15\%) seed the tool-catalogue table of one server from the BFCL v3 multi-turn function catalogue, which the dataset pool held as a public dataset; their trajectories are in the corpora of \S\ref{sec:setup} and were not removed, and the eight environments are withheld from the public release. MCP-Mark shares no server with the pool.

\bench shares no task, application or data with our environments, and Table~\ref{tab:overlap} measures the distance. We compare the request text of each of the 5{,}465 round-2 environments with the user prompt of each of the 600 public \bench tasks (their shared system prompt removed), after lower-casing, mapping digit runs to one placeholder and stripping punctuation. Lexical overlap is counted in shared word 13-grams and 8-grams; semantic proximity is the maximum TF-IDF cosine (sublinear tf, smoothed idf fitted on the union of the two corpora), with the benchmark's own leave-one-out self-similarity as the scale. No environment shares a 13-gram with any task. The six shared 8-grams are ISO-8601 timestamps and the phrase ``today is $\langle$date$\rangle$'', and none remains when digits are kept. 92.0\% of the benchmark tasks are closer to another benchmark task than to any of our requests; 23 of our requests (0.4\%) exceed the benchmark's median self-similarity and none exceeds its maximum. A dense embedding gives the same picture: 83.5\% of the benchmark tasks are closer to another benchmark task than to any of our requests; where half of our requests would exceed the benchmark's median self-similarity if they were drawn from its distribution, 0.5\% do, and none reaches its maximum, which two benchmark tasks that paraphrase each other across domains attain. Unrelated pairs score about 0.45 in this space, so only the comparison between the two columns carries meaning. The five nearest pairs share a business scenario, such as invoice approval, and nothing else. Our 223 servers are self-hosted open-source applications and the benchmark's 47 are simulated commercial SaaS tools; the two sets are disjoint, and the benchmark's worlds are synthetic. The benchmark places the governing rule in a record of the environment rather than in the prompt in 44.5\% of its tasks, so the request-level comparison is like for like.

\begin{table}[t]
\centering
\small
\setlength{\tabcolsep}{4pt}
\resizebox{\columnwidth}{!}{%
\begin{tabular}{@{}lrr@{}}
\toprule
Measure & ours vs.\ AB & AB vs.\ itself \\
\midrule
sharing a 13-gram & 0 / 5{,}465 (0.0\%) & 351 / 600 (58.5\%) \\
sharing an 8-gram & 81 / 5{,}465 (1.5\%)\textsuperscript{$\ast$} & 401 / 600 (66.8\%) \\
TF-IDF max cosine, median & 0.198 & 0.344 \\
TF-IDF max cosine, max & 0.442 & 0.680 \\
embedding max cosine, median & 0.719 & 0.779 \\
embedding max cosine, max & 0.930 & 0.965 \\
applications in common & 0 of 223 vs.\ 47 & --- \\
\bottomrule
\end{tabular}}
\caption{Overlap between our environments and \bench (AB). Rows 1--2: environments (left) or tasks (right) that share at least one word $n$-gram with some task; \textsuperscript{$\ast$}all six shared 8-grams are timestamps or ``today is $\langle$date$\rangle$'', and the count is 0 when digits are not normalised. Rows 3--6: for each benchmark task, the maximum cosine to our requests (left) and to the other 599 tasks (right), under TF-IDF and under a dense embedding (Qwen3-Embedding-4B).}
\label{tab:overlap}
\end{table}

\section{Corpus profile}
\label{app:corpus}

Figure~\ref{fig:corpus} profiles the 5{,}465 environments of the round-2 corpus by servers per environment, server frequency, rule archetypes and carriers, and operation kinds.

The round-2 corpus is 6{,}110 trajectories over 5{,}465 environments drawing on 223 distinct applications. A trajectory spans 2.51 applications on average (median 2; 1/2/3/4/5 applications in 31/3{,}361/2{,}359/315/44 trajectories), so \textbf{99.5\% of the corpus is cross-application}; the distribution is long-tailed but not degenerate, with the ten most frequent applications accounting for 60.6\% of application slots (58.8\% counted per environment, Figure~\ref{fig:corpus}b) and the most frequent single one for 15.2\%. Each task carries a verifier of 13.1 clauses on average (median 13, p90 24, max 54) and a gold witness of 12.2 actions (median 10, p90 25, max 64). The trajectories themselves run 14.8 assistant turns (median 13, p90 26, max 70) and 19{,}784 tokens (median 18{,}504, max 65{,}498); 33.3\% of the 120.9M tokens carry loss, and 23.4\% of trajectories contain at least one 4xx tool response, whose turns are masked out of the loss (Appendix~\ref{app:expdetails}).

\section{Every stage on the builder model}
\label{app:agents}

\begin{table}[!htb]
\centering
\small
\setlength{\tabcolsep}{3pt}
\resizebox{\columnwidth}{!}{%
\begin{tabular}{@{}lrrr@{}}
\toprule
Stage (agent on the builder model) & tried & admitted & $\geq$1/8 \\
\midrule
build, with rule atom      & 100 & 92 & 69/92 \\
build, without rule atom   & 100 & 98 & 78/98 \\
dispatch $\rightarrow$ build   & 30  & 30 & 14/20$^{\ast}$ \\
onboard $\rightarrow$ build    & 10  & 10 & 9/9$^{\ast}$ \\
instance repair            & 100 & 68 & 44/51$^{\ast}$ \\
\midrule
server repair (15 servers) & \multicolumn{3}{r}{188 of 203 recovered, 0 of 90 broken} \\
dispatch (skeletons)         & \multicolumn{3}{r}{100 of 100 valid, 100 new compositions} \\
onboarding (servers)        & \multicolumn{3}{r}{4 of 4 pass without intervention} \\
\bottomrule
\end{tabular}}
\caption{Each stage of \method run by an agent on DeepSeek-V4-Flash. \emph{Admitted}: passes the verifier correctness check. \emph{$\geq$1/8}: environments with at least one verifier-accepted trajectory in eight teacher attempts. $^{\ast}$Teacher attempts were run only on the environments that also passed a review of the request (20 of 30, 9 of 10, 51 of 68).}
\label{tab:agents}
\end{table}

Table~\ref{tab:agents} runs every stage of \method with an agent on DeepSeek-V4-Flash, about 100 cases per stage, against the existing artefacts as reference. The builder admitted 92 of 100 skeletons and the teacher solved 69 of them at least once (75.0\%, against 71.6\% for the main corpus). A repair agent given only the failure evidence and the original specifications of 15 servers, and none of the existing repairs, recovered 188 of the 203 environments that could not be built and broke none of the 90 that could. On 100 environments no teacher attempt solved, a diagnosis agent produced a valid diagnosis for 94 and repaired 68; 51 passed review and the teacher then solved 44. A dispatch agent given only the atom pools wrote 100 skeletons, all valid and none repeating a composition among the 3{,}254 reference skeletons that existed at the time of the run; an onboarding agent brought four new servers into the pool with no intervention, and in a second pass 12 of 14 servers that newly classified operation kinds asked for passed the onboarding check. A mining agent given only the vocabulary, tool metadata and dataset cards collected 1{,}825 candidate atoms from 296 public pages, and in a sample of 50 the cited page supported the atom directly in 42. The constraints of \S\ref{sec:recombine} each reject on their own: among 10{,}000 random skeletons that satisfy tool coverage, carrier hosting rejects 71.5\%, schema compatibility 41--45\% on a reconstructed dataset pool, and rule attachment 9.8\% when one intent must carry the rule (the reading of \S\ref{sec:recombine}) or 80.5\% when every intent must; together they admit 13--14\% under the one-intent reading and 1.6--1.7\% under the every-intent one.

\section{Where the content of related environments comes from}
\label{app:related}

Table~\ref{tab:related} reads the pipelines of \S\ref{sec:related} part by part: EnvScaler~\citep{envscaler}, AgentScaler~\citep{agentscaler}, TOUCAN~\citep{toucan}, Agent-World~\citep{agentworld}, AWM~\citep{awm}, ScaleEnv~\citep{scaleenv}, AgentSkiller~\citep{agentskiller}, EnvFactory~\citep{envfactory}, C-World~\citep{cworld}, AutoForge~\citep{autoforge}, COVERT~\citep{covert}, EnvCraft~\citep{envcraft}, CompoWorld~\citep{compoworld}. Every cell was checked against the full text of the paper. Two pipelines, TOUCAN and C-World, execute real remote services, so their tools and returned data are real, but they fix no initial state, cannot reset a task, and train by supervised fine-tuning only. EnvFactory and CompoWorld start from real tool interfaces (API documentation, crawled MCP specifications) and have an agent write the implementation, the schema and the data; EnvCraft starts from community scenario names and generates the rest. Agent-World mines part of its databases from the web and extends them with generated rows. AWM's released pipeline offers a code-only verifier beside its default code-plus-judge one; its authors report the code-only option as the weaker of the two, and it is the one the arm of Table~\ref{tab:baselines} uses.

\begin{table}[!htb]
\centering
\scriptsize
\setlength{\tabcolsep}{2.5pt}
\begin{tabular}{@{}lcccccccc@{}}
\toprule
& \multicolumn{4}{c}{source of each part} & & & \\
\cmidrule(lr){2-5}
Work & schema & tool & data & intent & cross-app & verifier & RL \\
\midrule
EnvScaler & \xmark & \xmark & \xmark & \xmark & \xmark & prog. & \cmark \\
AgentScaler & $\sim$ & $\sim$ & \xmark & \xmark & \xmark & prog. & \xmark \\
TOUCAN\textsuperscript{$\ddagger$} & --- & \cmark & \cmark & \xmark & \cmark & hybrid & \xmark \\
Agent-World & $\sim$ & \xmark & $\sim$\textsuperscript{$\S$} & \xmark & \xmark & hybrid & \cmark \\
AWM & \xmark & \xmark & \xmark & \xmark & \xmark & hybrid & \cmark \\
ScaleEnv & \xmark & \xmark & \xmark & \xmark & \xmark & prog. & \cmark \\
AgentSkiller & \xmark & \xmark & \xmark & \xmark & \cmark & hybrid & \xmark \\
EnvFactory & \xmark & $\sim$ & \xmark & \xmark & \cmark & prog. & \cmark \\
C-World\textsuperscript{$\ddagger$} & \cmark & \cmark & \cmark & \xmark & \cmark & hybrid & \xmark \\
AutoForge & \xmark & $\sim$ & \xmark & \xmark & \xmark & prog. & \cmark \\
COVERT & \xmark & \xmark & \xmark & \xmark & \xmark & hybrid & \cmark \\
EnvCraft & \xmark & \xmark & \xmark & \xmark & \cmark & prog. & \cmark \\
CompoWorld & $\sim$ & $\sim$ & \xmark & \xmark & \cmark & prog. & \cmark \\
\midrule
\method (ours) & \cmark & \cmark & \cmark & \cmark & \cmark & prog.\textsuperscript{$\dagger$} & \cmark \\
\bottomrule
\end{tabular}
\caption{Where the content of a synthesised agentic environment comes from. \cmark{} = derived from a real external artefact (source code, a dataset, a published template or procedure); $\sim$ = only names or interface descriptions are real, with behaviour or values written by an LLM; \xmark{} = written by an LLM. \emph{verifier}: what decides whether a trajectory solved its task; \emph{prog.} a program alone, \emph{hybrid} an LLM judge takes part in the decision (filters that discard tasks or trajectories are not counted). \textsuperscript{$\ddagger$}runs against live services, so schema, tools and returned data are real and no initial state is fixed; C-World's simulated mode is $\sim$/\xmark. \textsuperscript{$\S$}rows mined from the web, complexified with generated rows. \textsuperscript{$\dagger$}additionally tested: every verifier must reject the initial world and accept the reference solution before its environment is admitted (\S\ref{sec:admission}). }
\label{tab:related}
\end{table}

\section{Prompts}
\label{app:prompts}

Figure~\ref{fig:prompt-builder} gives the brief of the builder agent and Figure~\ref{fig:prompt-diagnosis} the prompt of the diagnosis agent, both cut and lightly edited to fit; cuts are marked \texttt{[...]}, machine paths are shortened, and code identifiers are kept and glossed in the captions. Both will be released with the code.

\begin{promptbox}{BUILDER AGENT --- \texttt{BUILDER.md}}
\begin{PromptVerbatim}
# BUILDER BRIEF — build one task pack from a skeleton spec

You are the builder. [...] You own ONE skeleton. Execute directly; do not spawn agents; the user is not watching. Report honestly (FAIL is fine). [...]

## Read first (once, in this order)
1. `SKILL.md` [...] (the builder's skill file).
2. The docstring plus the `World` and `Contract` sections of `packlib.py`.
3. `[...]` (two functions: `sim_gold`, `check`).
4. Two previously admitted environments as examples [...].
5. The data-atom cards [...].

## Your input: the skeleton spec
{SKELETON_JSON}
Field meaning:
- `apps`: the servers of the world [...].
- `deliverable_atoms`: what the task must produce, one atom per (app, kind). `pos` = how many positive checks the pack must carry for this atom, `neg` = how many negative checks. `check_kinds` says to what precision the pack must check this atom (the vocabulary [...]; e.g. `mail_sent_to_with_terms` → a mail to the right address whose body names the record and the amount; `row_cell_equals` → the exact cell value).
- `rule_atoms`: the business rule(s) the task turns on. `rule_shape` is an abstract description of the rule's shape (no names, no literal thresholds): instantiate it on your data, do not copy its wording into the goal. `has_exception` = the rule needs an override clause (hold / opt-out / already done / VIP …) and at least one distractor must trip exactly that clause. `archetype` is the rule's shape (threshold_routing / deadline_escalation / tiering / approval_chain / exclusion / reconciliation / budget_quota / compliance_flag); `carrier` says WHERE the rule text lives in the world: `policy_sheet` = a worksheet in the budibase workbook, `policy_mail` = an e-mail in the stalwart inbox, `notes_field` = a notes/description field on a CRM record, [...]. **The rule text must be literal and complete in the carrier, and must NOT be restated in the goal.** The goal only points at the carrier ("follow the payment rules the controller sent", "the workbook's policy worksheet has the requirements").
- `chain`: the abstract intent chain (read rule → select records → per-record actions → summary). Instantiate every step; do not add unrelated steps.
- `n_intents`: number of independently verifiable deliverables (reads/logins do not count). Keep it.
- `n_distractors_min`: at least this many records that LOOK eligible but fail exactly one clause of the rule; each gets its own NEGATIVE predicate ("the action did not happen for X": no mail to X, no row for X, X's cell unchanged).
- `predicate_budget`: positives + negatives must be >= this number.
- `delivery_shape`: per-record actions and/or one summary message.

## What to build (the skill's ten steps, adapted)
1. **tid** = a short name taken from the data (never the skeleton id). Claim it: `mkdir pack/<tid>` must succeed.
2. **Dataset**: pick from the data-atom pool [...]. Profile it with a few lines of python; the records, groups, thresholds, dates, recipients all come from real rows or statistics of them. Record `src=` on every seed call.
3. **World**: `W = World(apps, persona=(email, name), now=...)`; seed the record table/inbox/calendar/CRM; seed the **policy carrier** with the literal rule; add a little noise (`enron_mails`, `twcs_chat`, unrelated events); `spec = W.merge()` [...].
4. **Expected solution computed in code, never typed**: write the rule as a python function over the seeded rows, derive the matched records and the skipped ones, and `assert` the split (e.g. exactly 4 matches, >=2 distractors each failing a different clause, no ties).
5. **GOAL**: plain, natural English, 3–6 sentences; names the sheets/systems/channels it needs and the carrier of the rule; states dates it depends on ("today, <date>"); never lists the rules, the eligible records, the distractors, ids or recipients that the world already contains. Read it once as the agent: every value it needs must be findable by reading the world.
6. **Contract**: `C = Contract(GOAL, expected=...)`; positives from the computed expected (per matched record + summary), negatives per distractor and per invariant, all state-based [...]; `assert not C.seed_check(spec)`. **A predicate may only demand what the goal or the policy text explicitly asks for.** [...] Re-read every positive against the goal + policy text before writing it: a solver that follows both to the letter must pass it. [...]
7. **Gold steps** (the correct solution as tool calls, computed from the same expected): `steps = [(app, tool, args), ...]` then `bad, env = sim_gold(TID, spec, W, C, steps); assert not bad, bad`. [...]
8. `out, td = write_pack(TID, W, spec, C, atoms, skeleton="<skeleton_id>: <chain in one line>", notes_extra=...)` `atoms` = `[{"id": "a1", "kind": <deliverable kind>, "text": <one line>}, ...]` from the skeleton's deliverable_atoms.
9. `check(TID, "<skeleton_id>", dataset="<slug>", note="<<=80 chars>")` must print `CHECK PASS` (the admission check: every positive witnessed by a dropped write, every negative falsifiable, [...] resettable).
10. `README.md` in the pack dir (the skill's template), plus the rule text and where it lives.

Put all of it in ONE script `pack/<tid>/build.py` [...] and run it. Iterate on its asserts and on `check`'s report (<= 25 runs). When a
check fails: fix the contract, the world data or the gold steps. Only reword the goal if it was genuinely ambiguous, and say so in the
README.

[...]

## Newly onboarded apps [...]
[...] Each has a spec, a `register.py` (seed helpers, NOTES, LOGIN) and a `capability_card.json` (what it can hold: `content_entities`,
which `kinds` map to which write tools, `check_kinds`, `carriers`, `login`, `seed_helpers`). **Read the capability card and the
register.py docstring before seeding**: seed only the entities the card lists [...], sign in with the card's `login` tool, and realize
each deliverable atom with the tools the card lists for its `kind`. [...]

[...]

## Hard limits
- Write only under `pack/<tid>/` [...]. Everything else is read-only.
- Every whole-table read <= 12,000 chars, <= ~20 expected write calls, gold steps <= 30 calls.
- Seeded ids the agent passes back to tools must be integers; sheet column names without spaces and not `id`/`type`/`tableId`.
- No real personal e-mail addresses from datasets; use `<org>.example` addresses.

## Final message (<= 10 lines; nothing else is read)
tid, dataset slug, apps kept/dropped, rule archetype + carrier, n_intents, preds pos+neg, number of build.py runs, CHECK PASS/FAIL, and
<= 4 lines of findings (library gaps, spec bugs, what you could not express).
\end{PromptVerbatim}
\end{promptbox}
\nopagebreak
\captionof{figure}{Brief of the builder agent (\texttt{BUILDER.md}; cut and lightly edited to fit, cuts marked \texttt{[...]}). The skeleton is substituted for \texttt{\{SKELETON\_JSON\}}. The brief fixes the environment contract: the rule text lives only in its carrier, the expected solution is computed in code from the seeded rows, every positive assertion must be witnessed by a write and every negative must be falsifiable, and the environment is admitted only when \texttt{check} passes (\S\ref{sec:admission}). Code names: pack = environment; goal = request $q$; Contract, predicate, check = verifier $V$ and its assertions; gold steps, witness = reference solution $\pi$; tid = environment id; persona = the account the agent acts as; \texttt{sim\_gold} replays $\pi$ on $W_0$ and asserts \eqref{eq:seed}--\eqref{eq:final}; \texttt{check} is the admission check \eqref{eq:seed}--\eqref{eq:reset}.}
\label{fig:prompt-builder}
\medskip

\begin{promptbox}{DIAGNOSIS AGENT --- \texttt{AGENT\_BRIEF.md}}
\begin{PromptVerbatim}
# Instance diagnosis and repair

Read only the files in this case directory. `task.json` contains the request, verifier and reference solution (`witness`), `spec.json` contains its
world and tools, and `s0.json` through `s7.json` are eight failed teacher attempts. [...]

Determine the primary cause from the goal, world, recorded tool results and all eight trajectories. Write `diagnosis.json` with `tid`,
`category`, `cause`, and `action`. Choose exactly one category: `verifier_bug`, `task_data_environment_tool`, `task_data_or_tool`, or
`teacher_execution_failure`. `action` is `repair` or `no_change`. A teacher mistake is not evidence that the task needs weakening; use
`no_change` only when the package is sound and the failed attempts contain concrete execution or reasoning mistakes.

[...] Treat successful recorded API results as real
reads and mutations. For each strongest episode, map successful mutations and the persisted final state to the task's requested actions before
calling it an execution failure; a final verifier false can be a predicate false negative even when every requested mutation succeeded. [...]
Conversely, a reference solution that replays successfully proves reachability only: it may contain hidden constants that the public goal and
world never reveal. Do not use witness-only facts as requirements or as proof that the package is fair.

Before choosing a category, compare at least one strongest failed episode with the verifier clause by clause. Distinguish:

- a business-valid completion rejected by an underivable, contradictory, or misaddressed predicate: `verifier_bug`;
- a public source or entity promised by the goal but absent or inconsistent in the world/tool surface: `task_data_environment_tool`;
- an intrinsically ambiguous or impossible task/data/tool combination not localized to one environment artifact: `task_data_or_tool`;
- visible teacher mistakes despite a sound, fully discoverable package: `teacher_execution_failure`.

For `repair`, write `repaired_task.json` containing `goal_nl`, `env_id`, `verifier`, and a corrected `witness` (a list of `{tool,args}`
calls), and optionally `repaired_spec.json` if the world or tool definition must change. Preserve the business request. Change only the
smallest broken part; retain checks for requested actions and important exclusions. Never replace the verifier with a constant or make it
vacuously true. Do not copy a failed teacher sequence as a witness without checking it.

Audit positive coverage as well as false negatives. Enumerate every branch of the visible policy that selects an item and every action that
branch requires. [...]

For multi-branch policies, build a per-entity matrix before editing. [...] Recompute public formulas from their named operands exactly; do not
add a customary zero clamp, rounding rule, or display unit that the policy does not state.

For every selection and exclusion rule, identify each decision operand and its public source. [...] Do not infer the missing operand from the
witness. [...]

Audit every cross-system join. The entity identifier or mapping used to move from one system's record to another must be present in the public
task or world. Similar-looking identifiers in different namespaces are not a mapping. [...]

Treat a revision as a semantic diff. Before emitting it, list every verifier clause removed or weakened relative to the input candidate. [...]

Text checks should enforce semantic atoms rather than accidental surface form. [...]

When repairing an underivable exact timestamp, keep every public scheduling constraint and translate its wording exactly. `before X` requires
a strict `lt X`; `after X` requires `gt X`; `by X` or `no later than X` permits `lte X`; and `on or after X` permits `gte X`. [...]

When the goal requires one summary naming every acted-on item, [...] Honor the identifier form requested by the goal. [...] When the goal says
to leave every other item untouched, [...]

You may inspect the local runtime in `runtime/` and use `python3 check_candidate.py .` to replay your proposed witness, require a false
initial verifier and a true final verifier, and inspect errors. This is an offline smoke test. It does not establish business equivalence,
strict acceptance, or teacher success. Run it after each change. Record any unresolved issue in `diagnosis.json` rather than inventing a
successful repair. [...]

The files in this directory are the entire allowed input. Do not search [...] other cases. Do not alter `task.json`, `spec.json`, or the eight teacher episodes.
\end{PromptVerbatim}
\end{promptbox}
\nopagebreak
\captionof{figure}{\raggedright Prompt of the diagnosis agent (\texttt{AGENT\_BRIEF.md}; cut and lightly edited to fit, cuts marked \texttt{[...]}). The agent chooses one of four categories, \texttt{verifier\_bug}, \texttt{task\_data\_environment\_tool}, \texttt{task\_data\_or\_tool} and \texttt{teacher\_execution\_failure}, and one action, \texttt{repair} or \texttt{no\_change}; \texttt{verifier\_bug} is the verifier fault of \S\ref{sec:repair}, the two task/data categories are its world fault, and \texttt{teacher\_execution\_failure} its sound environment the teacher cannot solve. Code names: package = environment; witness = reference solution $\pi$; strict acceptance = the full admission check \eqref{eq:seed}--\eqref{eq:reset}, of which \texttt{check\_candidate.py} tests \eqref{eq:seed}--\eqref{eq:final}. It is the brief of the instance-repair run of Table~\ref{tab:agents}; the repairs of \S\ref{sec:repair} were made by an agent under an earlier iteration of this brief with the same four categories. The categories label the fault; which of the four pathways of \S\ref{sec:evolve} follows is decided from the root cause (Appendix~\ref{app:rootcauses}).}
\label{fig:prompt-diagnosis}
\medskip

\section{The 33 root causes}
\label{app:rootcauses}

\begin{figure*}[t]
\begin{codebox}{SERVER ATOM REPAIR --- \texttt{raneto} SPECIFICATION DIFF}
\begin{DiffVerbatim}
page_search  exec_spec.rank
-  {"method": "keyword", "fields": ["title", "body"]}
+  {"method": "substring"}
page_search  exec_spec.filter
+  {"op": "or", "clauses": [{"col": "title", "mode": "substring", "val": {"$arg": "search"}},
+                           {"col": "body", "mode": "substring", "val": {"$arg": "search"}}]}
page_edit  exec_spec.row_key        page_delete  exec_spec.lookup_by
-  {"filepath": {"$arg": "file"}}
+  {"slug": {"$arg": "file"}}
page_edit, page_delete  input_schema.file.description
-  "Page path relative to content_dir, e.g. 'tutorials/my-page'. Identifies the page to edit."  (edit)
-  "Page path relative to content_dir. Identifies the page to remove from the listing."  (delete)
+  "Page path relative to content_dir without the .md extension, i.e. the page slug,
+   e.g. 'tutorials/my-page'."
page_create  input_schema.category.required   - false              + true
page_create  guards.required                  - ["name"]           + ["category", "name"]
page_render  guards.requires_auth             - true               + false
page_render  exec_spec.eager                  + [{"entity": "PageMeta", "rel": "meta", "via": "filepath"}]
new tools                                     + category_list (Category)   + pagemeta_list (PageMeta)
\end{DiffVerbatim}
\end{codebox}
\caption{Repair of the raneto server atom: every field that differs between the original and the repaired specification, except the free-text description of \texttt{page\_render} (\texttt{-} before, \texttt{+} after).}
\label{fig:raneto}
\end{figure*}

\begin{table*}[t]
\centering
\scriptsize
\setlength{\tabcolsep}{3pt}
\begin{minipage}[t]{0.48\textwidth}\centering
\begin{tabular}[t]{@{}p{1.3cm}lr@{}}
\toprule
Layer & Root cause & Repairs \\
\midrule
\multirow{16}{1.25cm}{check-kind semantics (582)}
 & verifier\_wrong\_selection\_or\_arithmetic & 110 \\
 & verifier\_conflicts\_visible\_instruction & 86 \\
 & verifier\_unstated\_exact\_value & 81 \\
 & verifier\_format\_normalization & 63 \\
 & verifier\_equivalent\_expression & 42 \\
 & verifier\_overchecks\_linked\_text & 37 \\
 & verifier\_wrong\_storage\_location & 32 \\
 & verifier\_hidden\_schedule & 31 \\
 & verifier\_wrong\_entity\_id & 25 \\
 & verifier\_unsupported\_extra\_action & 14 \\
 & verifier\_negative\_substring & 14 \\
 & verifier\_hidden\_title\_or\_name & 14 \\
 & verifier\_cardinality\_or\_grouping & 10 \\
 & verifier\_hidden\_status\_or\_flag & 10 \\
 & verifier\_wrong\_record\_or\_field & 7 \\
 & verifier\_unpublished\_choice & 6 \\
\bottomrule
\end{tabular}
\end{minipage}\hfill
\begin{minipage}[t]{0.48\textwidth}\centering
\begin{tabular}[t]{@{}p{1.3cm}lr@{}}
\toprule
Layer & Root cause & Repairs \\
\midrule
\multirow{11}{1.25cm}{builder discipline (402)}
 & missing\_recipient\_or\_contact & 144 \\
 & missing\_source\_rows\_or\_fields & 140 \\
 & missing\_policy\_clause\_or\_exception & 46 \\
 & gold\_missing\_login & 26 \\
 & world\_seed\_integrity & 19 \\
 & ambiguous\_goal\_or\_role & 9 \\
 & reference\_answer\_misread\_visible\_data & 8 \\
 & cross\_app\_identity\_mapping & 6 \\
 & policy\_carrier\_conflict & 2 \\
 & builder\_field\_binding & 1 \\
 & gold\_wrong\_or\_missing\_action & 1 \\
\midrule
\multirow{4}{1.25cm}{service spec (8)}
 & tool\_argument\_contract & 5 \\
 & tool\_read\_or\_search\_defect & 1 \\
 & service\_relation\_write\_defect & 1 \\
 & service\_missing\_default\_or\_inferred\_field & 1 \\
\midrule
data card (4) & dataset\_semantic\_or\_unit\_gap & 4 \\
\midrule
pure instance (1) & runtime\_version\_mismatch & 1 \\
\bottomrule
\end{tabular}
\end{minipage}
\caption{The 33 root causes of the 997 repairs, by layer. 89 repairs also carry a secondary cause, which is not counted.}
\label{tab:rootcauses}
\end{table*}

Table~\ref{tab:rootcauses} lists the root causes. The 997 repairs fall into five layers: check-kind semantics 582, builder discipline 402, service spec 8, data card 4 and pure instance 1. A missing source fact, recipient or policy clause, a corrupt seed and a wrong reference action are assigned to builder discipline, because each is a builder contract failure that recurs; the one pure-instance repair is a runtime-version mismatch that affected one recorded episode. The first-pass labels of \S\ref{sec:repair} (695 verifier faults, 302 world faults) and these layers are different partitions: 173 of the 695 verifier-labelled repairs were adjudicated by the agent as builder discipline and 60 world-labelled ones as check-kind semantics.

\section{Repair examples}
\label{app:repairs}

\begin{table}[!htb]
\centering
\scriptsize
\setlength{\tabcolsep}{3pt}
\begin{tabular}{@{}lll@{}}
\toprule
Step & Kind added & Implemented by \\
\midrule
\multirow{4}{*}{31$\rightarrow$37}
 & add\_label & planka: add\_card\_label, create\_label \\
 & create\_metadata & paperless\_ngx: tag\_create \\
 & post\_batch & ledgersmb: batch\_post \\
 & record\_status\_change & akaunting: document\_sent, \ldots \\
\midrule
\multirow{4}{*}{37$\rightarrow$57}
 & reservation\_approve & librebooking: reservationseries\_approve \\
 & check\_pause & healthchecks: check\_pause \\
 & dataset\_publish & ckan: package\_create \\
 & available\_stock\_update & foodcoopshop: product\_edit\_quantity \\
\midrule
\multirow{4}{*}{57$\rightarrow$110}
 & comment\_or\_reply & 284 historical write steps \\
 & share\_or\_control\_access & 126 historical write steps \\
 & manage\_event\_program & 31 historical write steps \\
 & manage\_time\_tracking & 12 historical write steps \\
\bottomrule
\end{tabular}
\caption{Growth of the operation-kind vocabulary, with four of the kinds added at each step. Steps 1 and 2 give the server and tools that implement the kind; step 3 gives the number of historical write steps classified into it. The steps are vocabulary snapshots at onboarding, not evolution rounds.}
\label{tab:kindgrowth}
\end{table}

\begin{table*}[t]
\centering
\scriptsize
\setlength{\tabcolsep}{4pt}
\begin{tabular}{@{}p{1.3cm}>{\raggedright\arraybackslash}p{8.2cm}>{\raggedright\arraybackslash}p{5.6cm}@{}}
\toprule
Kind & URL & Atoms it yielded \\
\midrule
documentation & \texttt{docs.\allowbreak{}frappe.\allowbreak{}io/\allowbreak{}erpnext/\allowbreak{}auto-\allowbreak{}creation-\allowbreak{}of-\allowbreak{}material-\allowbreak{}request} & rule: \texttt{threshold\_routing}, \texttt{policy\_doc}; intents: \texttt{append\_row} on budibase, \texttt{send\_email} on stalwart \\
template & \texttt{n8n.\allowbreak{}io/\allowbreak{}workflows/\allowbreak{}16404-\allowbreak{}route-\allowbreak{}purchase-\allowbreak{}order-\allowbreak{}approvals-\allowbreak{}and-\allowbreak{}budget-\allowbreak{}alerts-\allowbreak{}with-\allowbreak{}gmail-\allowbreak{}slack-\allowbreak{}and-\allowbreak{}sheets} & rule: \texttt{budget\_quota}, \texttt{policy\_sheet}; intents: \texttt{send\_email} on stalwart, \texttt{create\_or\_update\_issue} on mantisbt, \texttt{append\_row} on budibase \\
procedure & \texttt{www.\allowbreak{}process.\allowbreak{}st/\allowbreak{}templates/\allowbreak{}accounts-\allowbreak{}payable-\allowbreak{}procedure-\allowbreak{}template/\allowbreak{}} & rule: \texttt{approval\_chain}, \texttt{notes\_field}; intents: \texttt{post\_batch} on ledgersmb, \texttt{record\_status\_change} on akaunting, \texttt{send\_dm} on zulip \\
procedure & \texttt{www.\allowbreak{}americanbar.\allowbreak{}org/\allowbreak{}groups/\allowbreak{}professional\_\allowbreak{}responsibility/\allowbreak{}publications/\allowbreak{}model\_\allowbreak{}rules\_\allowbreak{}of\_\allowbreak{}professional\_\allowbreak{}conduct/\allowbreak{}rule\_\allowbreak{}1\_\allowbreak{}7\_\allowbreak{}conflict\_\allowbreak{}of\_\allowbreak{}interest\_\allowbreak{}current\_\allowbreak{}clients/\allowbreak{}} & rule: \texttt{exclusion}, \texttt{policy\_doc}; intents: \texttt{update\_record} on espocrm, \texttt{create\_or\_update\_issue} on mantisbt \\
dataset & \texttt{www.\allowbreak{}kaggle.\allowbreak{}com/\allowbreak{}datasets/\allowbreak{}olistbr/\allowbreak{}brazilian-\allowbreak{}ecommerce} & domain crm; tables incl.\ \texttt{olist\_customers\_dataset} \\
dataset & \texttt{archive.\allowbreak{}ics.\allowbreak{}uci.\allowbreak{}edu/\allowbreak{}static/\allowbreak{}public/\allowbreak{}601/\allowbreak{}ai4i\%2B\allowbreak{}2020\%2B\allowbreak{}predictive\%2B\allowbreak{}maintenance\%2B\allowbreak{}dataset.\allowbreak{}zip} & domain manufacturing; numeric columns incl.\ \texttt{Torque [Nm]}, \texttt{Tool wear [min]} \\
dataset & \texttt{huggingface.\allowbreak{}co/\allowbreak{}datasets/\allowbreak{}theatticusproject/\allowbreak{}cuad/\allowbreak{}resolve/\allowbreak{}main/\allowbreak{}CUAD\_\allowbreak{}v1/\allowbreak{}master\_\allowbreak{}clauses.\allowbreak{}csv} & domain legal; columns incl.\ \texttt{clause\_type}, \texttt{clause\_text} \\
dataset & \texttt{huggingface.\allowbreak{}co/\allowbreak{}datasets/\allowbreak{}aryadevkota/\allowbreak{}dataco\_\allowbreak{}supplychain\_\allowbreak{}dataset} & domain commerce; columns incl.\ \texttt{Days For Shipping (Real)}, \texttt{Delivery Status} \\
repository & \texttt{github.\allowbreak{}com/\allowbreak{}ryanlelek/\allowbreak{}raneto} & server atom raneto (wiki), 12 tools after repair (Figure~\ref{fig:raneto}) \\
repository & \texttt{github.\allowbreak{}com/\allowbreak{}plankanban/\allowbreak{}planka} & server atom planka (boards), hosts the \texttt{card\_description} carrier \\
\bottomrule
\end{tabular}
\caption{Ten sources. Template, documentation and procedure rows give the rule atom (archetype, carrier) and the intents (kind on server) of the skeleton mined from the page; dataset rows give the dataset card's domain and columns; repository rows give the server atom compiled from the code.}
\label{tab:sources}
\end{table*}

\paragraph{A server atom.} Figure~\ref{fig:raneto} shows the repair of the wiki server raneto. Its search tool was a keyword search, but the runtime's keyword executor reads only the arguments \texttt{q} and \texttt{query} while the tool's argument is \texttt{search}, so every query returned an empty list. Its edit and delete tools located a page by \texttt{filepath}, whose seeded values carry the \texttt{.md} extension, while the argument description asks for a path without it, so a call made as described failed with ``page does not exist''. Categories and page metadata were readable by no tool. The repair makes search a substring filter over title and body, locates pages by slug, makes the page read public with its metadata, requires a category on page creation, and adds two list tools; the server goes from 10 to 12 tools.

\paragraph{An environment.} In \texttt{bozeman\_remodel}, the request allows a remodel work log per property, but the verifier required all four property names in one document. The repair accepts either one merged log or one log per property and keeps the requirement that every property is covered.

\paragraph{Operation kinds.} Two counts describe the operation kinds. The registered vocabulary grew with onboarding in three steps (Table~\ref{tab:kindgrowth}): 31 to 37 kinds with six servers (19 to 25), 37 to 57 with seven servers (25 to 32), and a final classification of all 3{,}935 historical write steps that kept a new kind only where no existing one fit, 110 in all. The order in which servers were onboarded was computed, not chosen: each candidate is ranked by how often mining named it before it was onboarded. The counts of Figure~\ref{fig:scaling}b, 81, 91 and 109, are the kinds observed in each round's writes and skeletons under that final taxonomy (81 counts the 1{,}166 of 1{,}204 round-0 environments the atomisation covers); round 1 adds ten. Kinds enter the vocabulary when a skeleton needs them: three mined government-domain skeletons named ckan's \texttt{dataset\_publish} and \texttt{dataset\_metadata\_patch}, which the validator rejected as \texttt{kind not in vocab}; onboarding ckan added both kinds and \texttt{resource\_publish}, and the skeletons then passed. The taxonomy was also corrected after the fact: a Bitwarden call \texttt{organizationuser\_update} with a \texttt{Type} argument, first classed as \texttt{manage\_membership\_or\_invitation}, was moved to \texttt{manage\_role\_or\_permission} because \texttt{Type} is the role field, while calls that change only \texttt{Status} stayed.

\section{Sources}
\label{app:sources}

The mined skeletons cite 556 distinct template, documentation and procedure URLs, most often from product documentation and published regulations (\texttt{docs.frappe.io} 76, \texttt{americanbar.org} 76, \texttt{docs.civicrm.org} 72); the dataset cards cite 163 distinct dataset URLs, and each of the 223 servers maps to one GitHub repository. Table~\ref{tab:sources} lists ten examples.

\end{document}